\documentclass{article}
\PassOptionsToPackage{table,HTML}{xcolor}

\usepackage{microtype}
\usepackage{graphicx}
\usepackage{subfigure}
\usepackage{booktabs} 

\usepackage{hyperref}

\usepackage[accepted]{icml2025}

\usepackage{amsmath}
\usepackage{amssymb}
\usepackage{mathtools}
\usepackage{amsthm}
\usepackage[capitalize,noabbrev]{cleveref}

\definecolor{blue_}{HTML}{bde0fe}
\definecolor{pink_}{HTML}{ffc8dd}

\theoremstyle{plain}

\theoremstyle{definition}

\theoremstyle{remark}

\usepackage[disable,textsize=tiny]{todonotes}

\usepackage[compact]{titlesec}
\usepackage{sidecap}
\titlespacing*{\section}{0pt}{*1.0}{*0.8}
\titlespacing*{\subsection}{0pt}{*0.5}{*0.3}
\titlespacing*{\subsubsection}{0pt}{*0.3}{*0.3}
\allowdisplaybreaks

\usepackage{hyperref}       
\usepackage{url}            
\usepackage{booktabs}       
\usepackage{amsfonts}       
\usepackage{nicefrac}       
\usepackage{microtype}      
\usepackage{xspace}
\usepackage{paralist}

\usepackage{wrapfig}
\usepackage{amsmath}
\usepackage{amssymb}
\usepackage{amsthm}
\usepackage{url}
\usepackage{paralist}
\usepackage{latexsym}
\usepackage{arydshln}
\usepackage{tabularx}
\usepackage{verbatim}
\usepackage{CJK}
\usepackage{flushend}
\usepackage{multicol}
\usepackage{multirow,makecell}
\usepackage{color}
\usepackage{colortbl,array}
\usepackage{silence}
\usepackage{arydshln} 
\usepackage{xspace}
\usepackage{soul}
\usepackage{pifont}

\usepackage{arydshln}

\usepackage{amsmath,amsfonts,bm}

\def\eqref#1{equation~\ref{#1}}

\def\1{\bm{1}}

\DeclareMathAlphabet{\mathsfit}{\encodingdefault}{\sfdefault}{m}{sl}
\SetMathAlphabet{\mathsfit}{bold}{\encodingdefault}{\sfdefault}{bx}{n}

\usepackage{tikz}

\makeatletter
\DeclareRobustCommand\onedot{\futurelet\@let@token\@onedot}
\def\@onedot{\ifx\@let@token.\else.\null\fi\xspace}

\def\ie{\textit{i.e}\onedot}

\makeatother

\newcommand{\hidethis}[1]{}

\definecolor{bblue}{HTML}{4F81BD}
\definecolor{oorange}{HTML}{F4C842}
\definecolor{rred}{HTML}{C0504D}
\definecolor{ggreen}{HTML}{9BBB59}
\definecolor{ppurple}{HTML}{9F4C7C}
\definecolor{darkgreen}{HTML}{228B22}
\definecolor{cred}{HTML}{D81B60}
\definecolor{cblue}{HTML}{1E88E5}
\definecolor{cyellow}{HTML}{FFC107}
\definecolor{nred}{HTML}{e41a1c}
\definecolor{nblue}{HTML}{377eb8}
\definecolor{ngreen}{HTML}{4daf4a}
\definecolor{lblue}{HTML}{6C8EBF}

\newlength\savewidth

\newcolumntype{x}[1]{>{\centering\arraybackslash}p{#1pt}}
\newcolumntype{y}[1]{>{\raggedright\arraybackslash}p{#1pt}}
\newcolumntype{z}[1]{>{\raggedleft\arraybackslash}p{#1pt}}

\newcommand{\app}{\raise.17ex\hbox{$\scriptstyle\sim$}}

\definecolor{deemph}{gray}{0.6}

\definecolor{baselinecolor}{gray}{.9}

\usepackage{color, colortbl}
\definecolor{emerald}{rgb}{0.31, 0.78, 0.47}
\definecolor{Gray}{gray}{0.9}
\definecolor{Highlight}{rgb}{0.89,0.89,0.94}
\usepackage[first=0,last=9]{lcg}

\definecolor{Highlight_blue}{RGB}{105, 245, 240}  
\definecolor{Highlight_pink}{rgb}{0.97,0.85,0.89}  

\newcommand{\chl}{\cellcolor{Highlight}}
\newcommand{\gthl}{\cellcolor{Gray}}

\renewcommand{\paragraph}[1]{\noindent\textbf{#1}}
\renewcommand{\bm}[1]{\mathbf{#1}}

\newcommand{\method}{\textrm{APM}\xspace}
\newcommand{\sbm}{\texttt{Seq\&BB} \text{Module}\xspace}
\newcommand{\scm}{\texttt{Sidechain} \text{Module}\xspace}
\newcommand{\rfm}{\texttt{Refine} \text{Module}\xspace}

\newcommand{\metric}[1]{\texttt{#1}}

\begin{document}

\twocolumn[

\icmltitle{An All-Atom Generative Model for Designing Protein Complexes}


\icmlsetsymbol{equal}{*}
\icmlsetsymbol{lead}{$\dagger$}
\icmlsetsymbol{core}{$\ddagger$}

\begin{icmlauthorlist}
~~~\icmlauthor{Ruizhe Chen}{equal,hnu,byted}~~~~~
\icmlauthor{Dongyu Xue}{equal,lead,byted}~~~~~~~
\icmlauthor{Xiangxin Zhou}{ucas,byted}

\icmlauthor{Zaixiang Zheng}{byted}
\icmlauthor{Xiangxiang Zeng}{hnu}
\icmlauthor{Quanquan Gu}{byted}
\end{icmlauthorlist}

\icmlaffiliation{hnu}{College of Computer Science and Electronic Engineering, Hunan University}
\icmlaffiliation{ucas}{School of Artificial Intelligence, University of Chinese Academy of Sciences}
\icmlaffiliation{byted}{ByteDance Seed~(this work was done during Ruizhe Chen and Xiangxin Zhou's internship at ByteDance Seed)}

\icmlcorrespondingauthor{Quanquan Gu}{quanquan.gu@bytedance.com}

\icmlkeywords{Machine Learning, ICML}

\vskip 0.3in
]



\printAffiliationsAndNotice{\icmlEqualContribution\ProjectLead} 


\begin{abstract}

Proteins typically exist in complexes, interacting with other proteins or biomolecules to perform their specific biological roles. 
Research on single-chain protein modeling has been extensively and deeply explored, with advancements seen in models like the series of ESM and AlphaFold2. 
Despite these developments, the study and modeling of multi-chain proteins remain largely uncharted, though they are vital for understanding biological functions. 
Recognizing the importance of these interactions, we introduce \method (All-\textbf{A}tom \textbf{P}rotein Generative \textbf{M}odel), a model specifically designed for modeling multi-chain proteins. 
By integrating atom-level information and leveraging data on multi-chain proteins, \method is capable of precisely modeling inter-chain interactions and designing protein complexes with binding capabilities from scratch. 
It also performs folding and inverse-folding tasks for multi-chain proteins.
Moreover, \method demonstrates versatility in downstream applications: it achieves enhanced performance through supervised fine-tuning (SFT) while also supporting zero-shot sampling in certain tasks, achieving state-of-the-art results.
We released our code at \url{https://github.com/bytedance/apm}.
    
\end{abstract}

\section{Introduction}
\label{sec:intro}

The application of AI technology in protein design has become a prominent research direction across biology, materials science, and artificial intelligence~\citep{notin2024machine}. 
The existing works can be categorized into two distinct approaches: general protein foundation models and protein design models for specific functions.
The former includes methods such as protein folding models~\citep{AlphaFold2,ESMFold,RoseTTAFold}, inverse-folding models~\citep{ProteinMPNN,ESM-IF1,LM-Design}, co-design models~\citep{ProtSeed,Multiflow}, and protein language models~\citep{ESMFold}.
These works are not specifically designed for any particular protein design task but aim to learn the general distribution of protein sequences and structures from extensive protein data. 
The latter approach focuses on the design of proteins with explicit biological activities, such as antibodies~\citep{MEAN}, binding peptides~\citep{PepFlow}, and enzymes~\citep{song2024generative}.

\begin{figure}[t]
\begin{center}
\centerline{\includegraphics[width=\columnwidth]{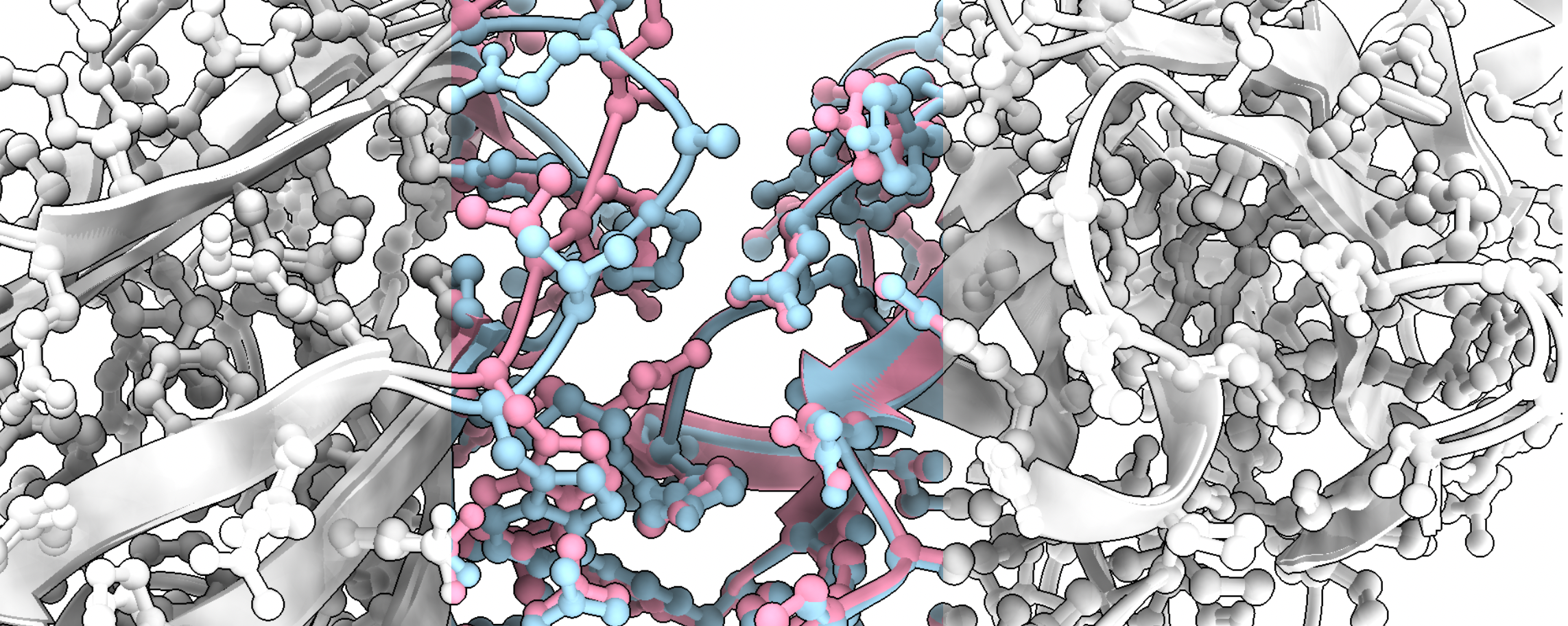}}
\vspace{-3mm}
\caption{Interactions cause minor atom-level protein structure changes in the binding surface (middle colored part). Blue indicates the isolated structure, pink indicates the binding structure.}
\label{fig:interaction}
\vspace{-3mm}
\end{center}
\end{figure}


General protein foundation models have demonstrated impressive performance across a broad range of tasks. However, these approaches focus solely on modeling single-chain proteins. 
In contrast, when dealing with proteins involved in specific functions, the target proteins usually appear in the form of complexes. 
Furthermore,
in multi-chain protein modeling, inter-chain interactions that occur at the atom-level play a crucial role (\cref{fig:interaction}). This necessitates incorporating models with atom-level information to enable precise learning of these interactions, which is fundamental for the effective modeling of multi-chain proteins.


To bridge this gap, we propose a novel method: \method~(All-Atom Protein Generative Model). \method facilitates the generation of multi-chain protein complexes with all-atom structures and can be applied to various tasks involving multi-chain protein complexes, including generation, folding, inverse-folding, and specific functional protein designs.
To develop such a generative model that can be used for designing bioactive complexes, 
we identify three core challenges: \textbf{multi-chain protein modeling}, \textbf{all-atom representation}, and \textbf{sequence-structure dependency}.

\textbf{Multi-Chain Protein Modeling.} Some efforts have attempted to adapt single-chain models for multi-chain protein tasks by using a poly-G pseudo sequence to connect different chains, treating them as a single chain, including AlphaFold2~\citep{AlphaFold2}, ESMFold~\citep{ESMFold}, and Linker-Tuning~\citep{Linker-Tuning}. 
This enables compatibility with multi-chain data but constrains the structural connectivity to head-to-tail linking, which is not representative of natural complex formations. 
In this work, we adopt a native method for modeling multi-chain proteins through both data integration and modeling strategies. 
For data, we use a mixture of single and multi-chain data in the training of \method. We believe that intra-chain modeling will benefit from the extensive amount of single-chain data. 
For modeling, our efforts include improving the model design and introducing conditional generation tasks. Key changes in model design
focus on encoding more information without altering the overall model structure, such as introducing inter-chain or intra-chain attention, thereby maintaining consistency across single and multi-chain proteins.

\textbf{All-Atom Representation.} In protein design with all-atom structures, the fundamental challenge lies in how to effectively represent atomic structures as different amino acid types have distinct atomic types, numbers, and basic structures. 
When the protein sequence is not determined, the representation of its atom-level structure directly influences the modeling approach. 
\citet{chu2024all} utilized an ensemble-based method to model the sidechain coordinates of all amino acid types simultaneously. 
\citet{AbDiffuser} represented sidechain structures by merging non-rotatable atoms into virtual atoms. 
\citet{P-all-atom} followed the method used in AlphaFold3~\citep{AlphaFold3} to model all-atom coordinates directly. 
We choose to enhance residue-level information with the sidechain for all-atom protein representation that includes amino acid type, backbone structure, and the sidechain conformation parameterized by four torsion angles.
This approach maintains computational efficiency while supplying atom-level information for modeling inter-chain interactions.

\textbf{Sequence-Structure Dependency.} 
The strong dependency between protein sequence and structure is the foundation for the success of folding
and inverse-folding models.
However, in the joint generation of protein structure and sequence, this dependency is disrupted during the independent noising process of each modality. 
This issue hampers effective learning of the dependency between sequence and structure. 
In \method, two strategies are implemented to enhance the dependency between the sequence and structure modalities. 
First, we decoupled the noising process for sequences and structures so that the noising level for each modality does not completely align, minimizing disruption of their dependency. 
Second, 
there is a 50\% probability of performing a folding/inverse-folding task, 
compelling the model to learn the dependencies from both directions.

Finally, \method has demonstrated its capability in modeling multi-chain proteins and generating bioactive complexes. 
It achieved state-of-the-art (SOTA) performance in antibody design and binding peptide design. 
Besides, \method also exhibited exceptional performance in conventional single-chain protein-related tasks.

We highlight our main contributions as follows:
\begin{compactitem}
\item \method natively supports the modeling of multi-chain proteins without the need to use pseudo sequence to connect different chains;
\item \method generates proteins with all-atom structures efficiently by utilizing an innovative integrated model structure; 
\item Experiments related to general protein demonstrate that \method is capable of generating tightly binding protein complexes, as well as performing multi-chain protein folding and inverse folding tasks;
\item Experiments in specific functional protein design tasks show that \method outperforms the SOTA baselines in antibody and peptide design with higher binding affinity.
\end{compactitem}

\begin{figure*}[ht]
\begin{center}
\centerline{\includegraphics[width=1.0\textwidth]{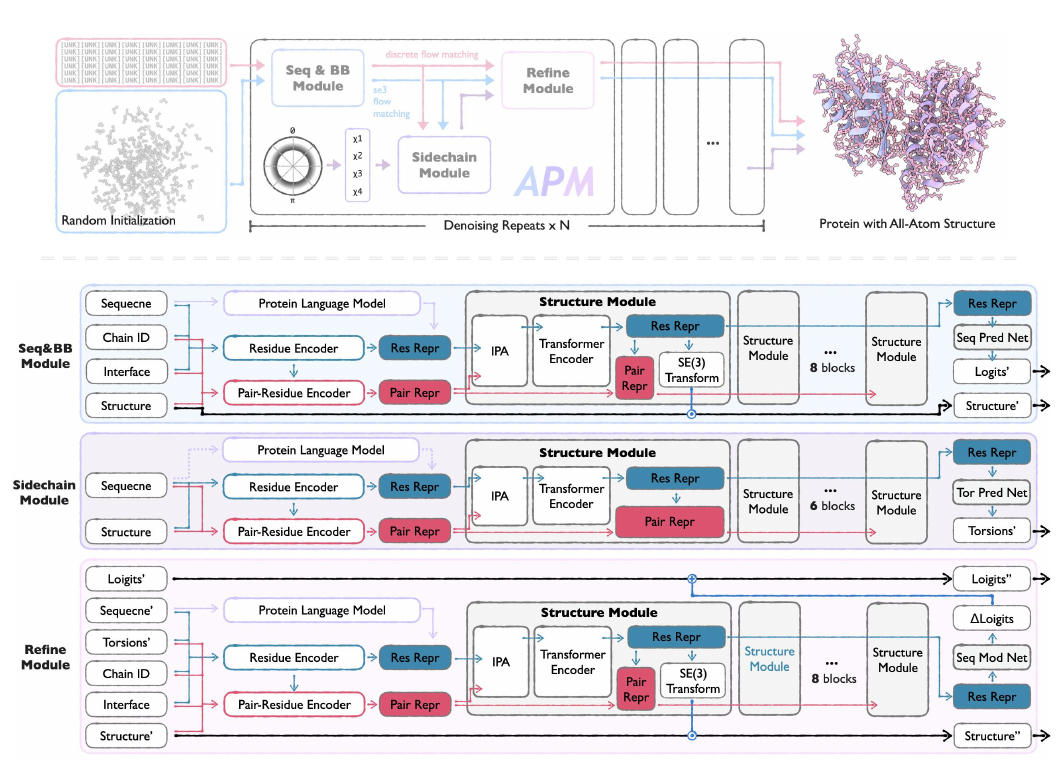}}
\vspace{-3mm}
\caption{Overview of \method. \method consists of three modules: (1) A flow-matching based \texttt{Seq\&BB} Module for generating backbone structure and sequence simultaneously; (2) a \texttt{Sidechain} Module for generating the all-atom structure based on the previous module's generation; (3) A \texttt{Refine} Module adjusts the sequence and structure with all-atom information. The iterative denoising process enables the generation of multi-chain proteins with all-atom structure. The detailed architecture of each module are presented below.}
\label{fig:illustration}
\vspace{-8mm}
\end{center}
\end{figure*}


\section{Related Work}
\label{sec:related}
\paragraph{Protein Foundation Models.} 
The breakthrough achievements in protein structure prediction, marked by AlphaFold series (AlphaFold1-3~\citep{AlphaFold1,AlphaFold2,AlphaFold3} and RoseTTAFold~\citep{RoseTTAFold,RoseTTAFold-All-Atom}, have revolutionized the field of protein science. 
With these developments, protein language models~\citep{ESM,Progen} have emerged as powerful tools. 
The series of ESM~\citep{ESM,ESMFold,ESM3}, trained on large-scale protein sequence data, have demonstrated remarkable capabilities in protein understanding and generation. 
Meanwhile, certain methods considered protein design workflow in two stages:  RFdiffusion~\citep{RFdiffusion} tackles backbone structure generation using diffusion models, while ProteinMPNN~\citep{ProteinMPNN} specializes in sequence design through message-passing neural networks. FrameFlow~\citep{yim2024improved,yim2023fast} and FoldFlow~\citep{FoldFlow1} present developments applying SE(3) flow matching approaches to protein structure generation. 
FoldFlow2~\citep{FoldFlow2} further demonstrates the integration of protein language models for structure generation.
Besides, Chroma~\citep{Chroma2023} introduces a unified approach to protein design through a generative model that can directly sample novel protein structures and sequences while being conditioned to target specific properties and functions
The field has also seen approaches like Multiflow~\citep{Multiflow}, ProteinGenerator ~\citep{ProteinGenerator}, and Protpardelle~\citep{Protpardelle}, which enable generation of both sequence and structure. 
More recently, SaProt~\citep{SaProt,Saprothub,ProTrek} and DPLM series~\citep{LM-Design,DPLM,DPLM-2,hsieh2025dplm21} have further advanced protein token modeling by incorporating structural information into the pre-training process, enabling better understanding of protein sequences.

\paragraph{Functional Protein Design.} 
Target-specific protein design has made remarkable advances recently.
In antibody design, approaches like HERN~\citep{HERN}, DiffAb~\citep{DiffAb}, MEAN~\citep{MEAN}, and dyMEAN~\citep{dyMEAN} have demonstrated the ability to generate functional antibodies.
Besides, ~\citet{HTP}, ~\citet{AbX}, and~\citet{AbGNN} introduced pre-trained protein language models as sequence priors to improve antibody design. 
For peptide design, methods such as PPFlow~\citep{PPFlow}, PepFlow~\citep{PepFlow}, PepGLAD~\citep{PepGLAD} and CpSDE~\citep{zhou2025cpSDE} focus on designing bioactive peptides. 



\section{APM}
\label{sec: method}
In this section, we present \method, an all-atom generative protein model for designing bioactive complexes with the all-atom structure. We first define how we represent the all-atom structure in \cref{method:full_atom_repr}. Then we introduce the model architecture of \method in \cref{method:model_architecture}, and in \cref{method:training}, we introduce the learning objective and training process. Finally, we introduce the sampling method in  \cref{method:sampling}.

\subsection{Representation for Protein All-Atom Structure}
\label{method:full_atom_repr}


In this study, we divide the goal of our approach into two parts: \textbf{1}, \textbf{foundational modeling of intra-chain sequences and structures} (determining what constitutes a plausible protein sequence and structure); \textbf{2}, \textbf{modeling of inter-chain interactions} (understanding how proteins interact with each other). Residue-level information is generally sufficient for intra-chain modeling of protein sequences and structures. Methods such as AlphaFold2~\citep{AlphaFold2}, ESMFold~\citep{ESMFold}, and ProteinMPNN~\citep{ProteinMPNN}, which leverage residue-level information, have demonstrated high-quality protein structure modeling. 
However, these models often require an additional relaxation step to resolve atomic clashes on the sidechains due to the lack of finer detail modeling. Therefore, we enhance residue-level information by incorporating sidechain conformations into the protein representation.

While modeling the all-atom coordinates provides the most detailed view of interactions, it considerably raises complexity and limits the ability to model longer proteins, especially for multi-chain proteins.
Sidechains of amino acids are not entirely free in their structures but have some conformations.
Each amino acid has up to four rotatable bonds in its sidechain while maintaining a largely consistent atomic structure between these bonds.
Therefore, combining amino acid type with sidechain torsion angles offers a comprehensive representation of sidechain conformation.

Ultimately, we adopted a representation that includes \textbf{amino acid type}, \textbf{backbone structure}, and \textbf{sidechain torsion angles}.
This approach maintains computational efficiency while providing richer information for modeling inter-chain interactions.

\begin{figure*}[t]
\begin{center}
\centerline{\includegraphics[width=1.0\textwidth]{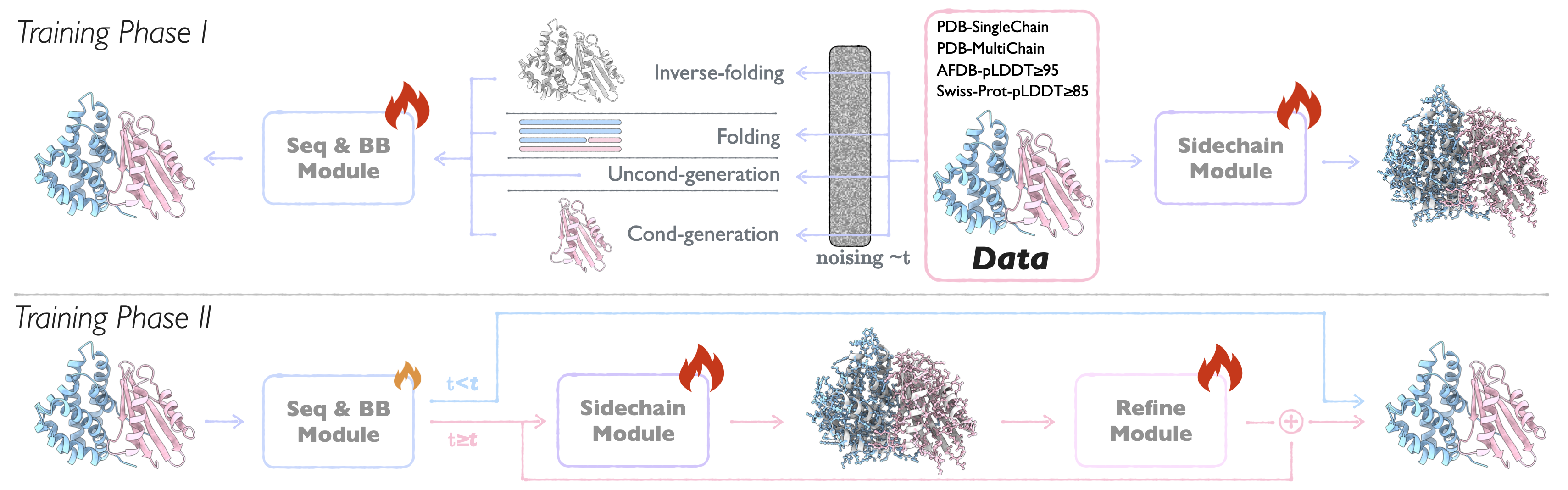}}
\vspace{-2mm}
\caption{The two-phase of the training process of \method. In training phase I, the \sbm and \scm are trained separately. In training phase II, the three modules form the integral \method, and are trained in an iterative paradigm. In any phase, the training data is a mixture of PDB~\citep{PDB} single/multi-chain proteins, Swiss-Prot proteins, and AFDB~\citep{AFDB} proteins.}
\label{fig:training}
\vspace{-8mm}
\end{center}
\end{figure*}

\paragraph{Notations.} 
The all-atom protein structure is represented as a collection of amino acid types, backbone frames, and sidechain torsion angles~\citep{AlphaFold2,lehninger2005lehninger}. 
A multi-chain protein complex $\mathcal{P}$ is composed of $K$ chains and $N=\sum_k k_N$ residues in total.
For the $k$-th chain, the amino acid sequence is denoted as $\bm{S}_k = [S_k^{k_1}, S_k^{k_2}, \ldots, S_k^{k_N}]$, where $S_k^{k_i} \in \mathcal{A}$ and $\mathcal{A}$ is the set of 20 standard amino acids.
Meanwhile, the backbone structure of this chain is characterized by rigid frames $\bm{T}_k =\left[T_k^{k_1}, T_k^{k_2}, \ldots, T_k^{k_N}\right]$, where each $T_k^{k_i} \in \mathrm{SE}(3)$ consists of a rotation $R_k^{k_i} \in \mathrm{SO}(3)$ and a translation vector $\mathbf{x}_k^{k_i} \in \mathbb{R}^3$, mapping rigid transformations from ideal peptide geometry~\citep{engh2006structure}.
The sidechain torsion angles are denoted as $\bm{\chi}_k=[\chi_k^{k_1}, \chi_k^{k_2}, \ldots, \chi^{k_N}]$, where $\chi_k^{k_i} \in[0, 2\pi)^{4}$ corresponds to the torsions of rotatable bonds in the sidechain of $i$-th residue.
For \textbf{brevity}, we slightly abuse the notation such that $\bm{S} = \bigcup_{k} \bm{S}_k$, $\bm{T} = \bigcup_{k} \bm{T}_k$, $\bm{\chi} = \bigcup_{k} \bm{\chi}_k$, where chain indices (\ie, $k$) are omitted hereafter unless needed.


\subsection{An Integrated Architectural Design of APM}
\label{method:model_architecture}

\subsubsection{Overall Architecture}

To implement an All-Atom Protein Generative Model, we designed the \method consisting of three distinct modules: the \sbm, the \scm, and the \rfm (\cref{fig:illustration}). The \sbm is a flow-matching-based protein generative model that handles the co-generation of sequence and structure at the residue level. The \scm serves as an all-atom completion model, predicting the sidechain conformations for proteins generated by the \sbm. The \rfm is an All-Atom Protein Refinement model, refining the generated proteins to make them more akin to natural proteins while resolving structural clashes.

When using \method for protein generation, the final result is progressively generated
from noise to data, with timestep $t$ from 0 to 1.
Notably, the \scm and \rfm are activated only after timestep $t\ge\mathcal{T}$~($\mathcal{T}=0.8$ here).
We believe that the quality of proteins produced by the \scm with $t$ far from data time (1) is insufficient to support high-quality predictions by the \scm and causes the meaningless in the refinement by the \rfm.

The motivation for designing this integrated architecture, rather than using a single model to directly generate proteins with all-atom structures lies in the incompatibility of training the sidechain prediction model with the sequence and structure flow-matching model. Two key reasons prevent us from doing this:
\textbf{(1)} sidechain prediction requires real sequences and structures to obtain accurate sidechain conformation labels, while the flow-matching model uses noised sequences and structures as input;
\textbf{(2)} while it is possible to generate sidechain conformations using a flow-matching approach rather than a packing model (one-step prediction), this would require providing a noised sidechain conformation, $\chi_t$, which still contains amino acid type information, \textbf{leading to sequence information leakage} (details in \cref{appendix:tor_distribution}). 
This is evidenced by the rapid convergence of amino acid type loss during training. During sampling, the absence of truly noised torsion angles, $\chi_t$, significantly degrades the model’s performance in the inference phase.

With the generation of sequence \& backbone structures separated from sidechain conformations, an additional module is necessary to allow all-atom information to influence the design of the backbone structure accordingly. For this purpose, we developed the third module of \method, the \rfm. It receives outputs from both the \sbm and \scm, making it all-atom aware. Based on this comprehensive information, the \rfm further optimizes the sequence and backbone structure to ensure the overall structure more closely resembles natural proteins.

\subsubsection{Sub-module Architectures}

The core structure of the three submodules is essentially the same. We use stacked structure modules derived from AlphaFold2 as the trunk of \method. Each structure module is composed of IPA~\citep{AlphaFold2} and a Transformer Encoder, which is employed to update residue information and pair-residue information. The differences among the submodules lie in the encoding of the input and the distinctions in the output, driven by the various modeling tasks. Apart from this, the \sbm and the \rfm maintain consistent model sizes, whereas the \scm has fewer structure module blocks and a smaller hidden dimension. We believe that predicting sidechain conformations is a relatively simple task, and utilizing a smaller model for enhancing efficiency.

\subsubsection{Integration of Protein Language Model}

A robust understanding of protein sequences requires a large-scale model trained on tens of millions of sequence data~(like the 3B-parameters ESM2 in ESMFold, which is trained on 65 million unique sequences, is responsible for sequence understanding), or alternatively, using MSA as the sequence representation (like AlphaFold2/3). \method is trained on protein data with structural information, yet the available volume of such data is not sufficient to support learning the intricacies of protein sequence understanding.
To address this, we integrated protein language models (PLMs) into all modules to enhance protein sequence understanding. 

We utilized ESM2-650M, the widely adopted protein language model,
to represent the input sequences. Drawing from ESMFold's approach, we used learnable weights to aggregate the representations from each layer of the protein language model, yielding the final amino acid encodings.
It's important to note that ESM2 is only trained on single-chain data. Therefore, when encoding multi-chain proteins with ESM2, each chain is encoded individually.

\subsection{Training of \method}
\label{method:training}
In order to train \method with the integrated architecture, we designed a two-phase training approach (\cref{fig:training}). In phase I, \sbm and \scm are trained separately. In phase II, the three modules are joint-trained in an iterative paradigm. All the details refer to \cref{appendix:training_details}. 

\subsubsection{Training of \sbm}

\sbm is the foundation model in \method to generate the sequence and backbone structure trained in a flow-matching manner with tasks of unconditional generation, conditional generation, folding, and inverse folding.
The primary learning objective is reconstructing either sequence or structure, or both, from a noisy state.
As we decouple the noising processes of the two modalities, we denote the noised sequence as \( \bm{S}_{t_S} \) and the noised structure as \( \bm{T}_{t_T} \), where \( t_S, t_T \sim \mathcal{U}(0,1) \) represent intermediate time steps. 
We also denote the original sequence and structure as \( \bm{S}_1 \) and \( \bm{T}_1 \).
Then the learning objectives of \sbm are: $p_\texttt{Seq\&BB}(\bm{S}_1, \bm{T}_1|\bm{S}_{t_S},\bm{T}_{t_T},t_S,t_T)$ for unconditional generation; $p_\texttt{Seq\&BB}(\bm{S}_1|\bm{S}_{t_S},\bm{T}_{1},t_S)$ for inverse-folding; $p_\texttt{Seq\&BB}(\bm{T}_1|\bm{S}_{1},\bm{T}_{t_T},t_T)$ for folding.

For each of the three tasks, there is a conditional version for multi-chain data, in which part of the target modalities is set as the noiseless state.
The flow-matching loss $\mathcal{L}_\text{flow-matching}$ is defined over sequence and backbone structure as:
\begin{align*}
    \mathcal{L}_\text{flow-matching} = \mathcal{L}_{\text{discrete}} + \mathcal{L}_{\text{SE}(3)}
\end{align*}
It consists of two components.
For the sequence, $\mathcal{L}_{\text{discrete}}$ measures the cross-entropy between the predicted sequence distribution $p_\texttt{Seq\&BB}(\hat{\bm{S}}_1 | \bm{S}_{t_S}, \bm{T}_{t_T})$ against the true sequence $\bm{S}_1$.
For the structure, it is the mean squared error between the vector fields calculated from the noisy structure $\bm{T}_{t_T}$ to the generated structure $\hat{\bm{T}}_1$ and to the true structure $\bm{T}_1$. For the complete mathematical formulation, refer to \cref{appendix:model}.

The refinement by the \rfm can be considered as a posterior correction, where the generated protein $(\hat{\bm{S}}_1,\hat{\bm{T}}_1)$ at each sampling step is corrected to $(\tilde{\bm{S}}_1,\tilde{\bm{T}}_1)$ before being noised for the next step. 
We hope to ensure that the direction of correction at each step is as consistent as possible. 
By achieving this, the corrections at each step can accumulate, leading to improved performance. 
This consistency necessitates that the predicted $(\hat{\bm{S}}_1,\hat{\bm{T}}_1)$ at each $(t_S,t_T)$ is aligned, which in turn requires the \sbm to maintain a smoother generative trajectory. 
To facilitate this, we incorporated a consistency loss, $\mathcal{L}_\text{consistency}$, into the \sbm to minimize the variations between predictions for adjacent $t$ (details in \cref{appendix:training_details:bb_model}).

The final training loss of \sbm is defined as:
\begin{align*}
\mathcal{L}_\texttt{Seq\&BB} = \mathcal{L}_\text{flow-matching} + 0.3 \times \mathcal{L}_\text{consistency}
\end{align*}
The training loss is consistent in two phases, the only difference is the learning rate.

\subsubsection{Training of \scm}

The learning objective of \scm is to predict the sidechain torsions, $\bm{\chi}$, given a protein with sequence and backbone structure. In the training phase~I, the learning objective is \textbf{packing}, $p_\text{Sidechain}(\bm{\chi}|\bm{S}_1,\bm{T}_1)$, which is no different from the normal packing model. While in phase II, the learning objective is switched to $p_\text{Sidechain}(\bm{\chi}|\hat{\bm{S}}_1,\hat{\bm{T}}_1)$, which means the \textbf{reconstruction} of the ground truth sidechain from the predicted $(\hat{\bm{S}}_1,\hat{\bm{T}}_1)$. Besides, we also want \scm to keep the ability of packing, so there is a 50\% probability that packing will continue to be used as the learning objective in phase II.

Training loss of \scm is also different for each learning objective. For \textbf{packing}, the loss consists of supervised torsion angle loss and all-atom Frame Aligned Point Error (FAPE) loss~\citep{AlphaFold2}, is defined as: 
\begin{align*}
\mathcal{L}_\text{Packing}=\mathcal{L}_{\chi}+\mathcal{L}_\text{FAPE}
\end{align*}
For \textbf{reconstruction}, we only maintain torsion angle loss, $\mathcal{L}_{\chi}$, as the input protein sequence and structure may not match the ground truth, resulting in the inappropriateness for calculating error on all frames (details in \cref{appendix:training_details:sc_model}).

\subsubsection{Training of \rfm}
The \rfm $p_\text{Refine}(\tilde{\bm{S}}_1, \tilde{\bm{T}}_1|\hat{\bm{S}}_1, \hat{\bm{T}}_1, t_S,t_T,\hat{\bm{\chi}})$ is tasked to predict the real protein based on the generated one, $(\hat{\bm{S}}_1, \hat{\bm{T}}_1)$, with the all-atom level information formed with the predicted sidechain torsions $\hat{\bm{\chi}}$. 
Thus, we define a correction loss on the corrected protein $(\tilde{\bm{S}}_1, \tilde{\bm{T}}_1)$ as the learning objective as:
\begin{align*}
\mathcal{L}_{\text{corr}} = - \log p(\bm{S}_1 | \tilde{\bm{S}}_1) + \|\tilde{\mathbf{x}}_1 - \mathbf{x}_1\|^2 + \|\tilde{R}_1 - R_1 \|^2
\end{align*}


Besides, we also incorporate auxiliary objectives in the training of \rfm, including backbone FAPE loss, $\mathcal{L}_\text{BB-FAPE}$, and residue distogram prediction loss, $\mathcal{L}_\text{dist}$ (details in \cref{appendix:training_details:rf_model}). Finally, the training loss of \rfm is defined as: $\mathcal{L}_\text{Refine}=\mathcal{L}_\text{corr}+0.25\times\mathcal{L}_\text{BB-FAPE}+0.25\times\mathcal{L}_\text{dist}$.

\subsubsection{Training in Phase II}

In the second phase of the training, we did not train the three modules simultaneously as each module requires a distinct $t$ range.
Instead, we employed an iterative approach for training these three modules. Given that the modules of \texttt{Seq\&BB} and \texttt{Sidechain} were already trained in phase I, they are only trained for 2 steps in each iteration in phase II, whereas the Refine module requires 8 steps of training. Each cycle comprises 12 steps, with the steps of 2-2-8 respectively.

The ultimate training loss of \method\ is defined as the expectation over timesteps and each residue in the protein:
\begin{align*}
\mathcal{L}=\mathbb{E}_{t\sim\mathcal{U}[0,1]}\left[
\mathcal{L}_\text{Seq\&BB} +
\mathcal{L}_\text{Packing} + 
\mathcal{L}_\text{Refine}
\right]
\end{align*}



\subsection{Sampling Strategy}
\label{method:sampling}

For structure sampling, we directly use the structure predicted by \sbm or the \rfm corrected one as the model output if it is activated.

For sequence sampling, the strategy is different.
To fully leverage the Protein Language Model, we update all the residues at each inference step and only keep the residues located in the positions with top $\text{max}(\log(\text{prob}))$. For the top K positions (where K is the number of amino acids to be unmasked at the current $t$), we sample the amino acid types based on the corresponding logits with a carefully designed strategy composed of temperature annealing sampling and $\arg\max$. The remaining positions are set to \texttt{[MASK]} token (details refer to \cref{appendix:experimental_details:seq_sampling}). 
This decoding strategy also led us to abandon the flow-matching training approach for the \scm, as the amino acid type at each position may change during the sampling process, making it inappropriate for the \scm to rely on the sequence from the previous step for prediction.

\section{Experiments}
\label{sec:experiments}

\subsection{Data Curation}


\textbf{Single-chain data} is built from three sources: PDB~\citep{PDB}, Swiss-Prot~\citep{Swiss-Prot}, and AFDB~\citep{AFDB}. For PDB samples, we followed the data processing flow in MultiFlow, resulting in 18684 samples. For Swiss-Prot samples, we selected the samples with a pLDDT~\citep{AlphaFold2,lDDT} greater than 85, resulting in 140769 samples. For AFDB samples, we take a more rigorous filter, leaving samples with a pLDDT greater than 95, which resulted in 28041 samples. Finally, we got 187494 single-chain samples.

\textbf{Multi-chain data }is built from PDB Biological Assemblies~\citep{rose2016rcsb}. To prevent potential information leakage in downstream
tasks, we discarded samples that met any of the following conditions: (1) the sample's PDB ID is present in SAbDab~\citep{SAbDab}; (2) the sample contains at least one chain with length less than 30, which is considered a peptide~\citep{PepGLAD,LNR}. The last condition led to the removal of a substantial number of samples (12,163).
In many cases, the peptides played crucial roles in stabilizing the complexes and only removing peptides is unreasonable.
Consequently, we opted to exclude this subset of data entirely from training. We also removed samples with lengths exceeding 2048
or lacking cluster IDs. Finally, we got 11620 multi-chain samples.

\textbf{Cropping. }During the training process, we performed random cropping~\citep{AlphaFold-Multimer} on multi-chain samples with residues exceeding 384 to prevent out-of-memory. Cropping was centered around the randomly selected inter-chain residue pair at the binding interface, retaining the 384 amino acids nearest to the pair. PLM encodes the sequence of cropped samples before cropping.

\subsection{Single-Chain Protein Related Tasks}

\label{exp:single_chain}
While \method is specifically designed for modeling multi-chain proteins, 
it also possesses the capabilities of those foundation models designed for single-chain proteins, including folding and inverse-folding. We validated the folding and inverse-folding capabilities of \method on a PDB date split used by MultiFlow. We compared it with specialized models, including ESMFold and PorteinMPNN, as well as co-design models capable of performing multiple tasks, including ESM3 and MultiFlow*(without distillation). We utilize \metric{RMSD} and \metric{TMscore}~\citep{TM-score} between predicted and ground truth structures to evaluate folding performance, and self-consistency~\citep{self-consistency} \metric{TMscore} (\metric{scTM}), amino acid recovery (\metric{AAR}) and perplexity to evaluate inverse-folding performance. The perplexity (\metric{ppl}) is provided by ProGen2-base~\cite{Progen,Progen2}. The results are shown in \cref{tab:single_chain_folding}.

\method can also perform unconditional protein generation. Besides ESM3 and MultiFlow*, we compared two methods capable of all-atom design, ProteinGenerator and ProtPardelle. For this task, we followed the evaluation methods in ProteinBench~\citep{ProteinBench} and presented the average \metric{scRMSD} and \metric{scTM} for proteins with lengths of 100-300 in \cref{tab:single_chain_uncond}.
\method achieved competitive performance compared to other co-design methods in all three tasks.

\begin{table}[t]
    \fontsize{7.2}{9}\selectfont
    \tabcolsep 2.5pt
    \centering
    \caption{Performance comparison of protein folding (blue highlighted) and inverse-folding tasks (pink highlighted). For each metric, we report the average/median performance.}
    \label{tab:single_chain_folding}
    \begin{tabular}{l|ccccc}
\toprule
Method &  \cellcolor{blue_} RMSD $\downarrow$  & \cellcolor{blue_} TM $\uparrow$ & \cellcolor{pink_} scTM $\uparrow$ & \cellcolor{pink_} AAR(\%) $\uparrow$ & \cellcolor{pink_} ppl $\downarrow$ \\
\midrule
ESMFold    &  2.84/1.19      &     0.93/0.97   &   -   &   -  & -  \\
ProteinMPNN      &   -    &   -  &     0.94/0.97     &      46.58/46.76  & 11.44/11.48   \\
\midrule
ESM3(1.4B)       &   \textbf{4.71}/\textbf{2.27}      &   0.83/\textbf{0.91}      &    0.94/0.97      &      49.50/49.42  & \textbf{8.64/7.90} \\
MultiFlow*  &   15.64/16.08    &   0.53/0.49     &    0.94/0.96      &    37.74/37.59    & 10.86/10.94   \\
\midrule
\chl \method & \chl  4.83/2.64     &  \chl \textbf{0.86}/\textbf{0.91}      &  \chl  0.94/0.97       &  \chl  \textbf{50.44}/\textbf{50.41}   &  \chl 8.74/8.10  \\
\bottomrule
\end{tabular}
\end{table}

\begin{table}[t]
    \fontsize{7.5}{9}\selectfont
    \tabcolsep 3.0pt
    \renewcommand{\arraystretch}{1.0}
    \caption{Performance comparison of different methods for various protein lengths. We evaluate the methods on three different length ranges (100, 200, 300) using \metric{scTM} and \metric{scRMSD}.}
    \label{tab:single_chain_uncond}
    \centering
    \begin{tabular}{l|cc|cc|cc}
\toprule
\multirow{2}{*}{Method} & \multicolumn{2}{c|}{Length 100} & \multicolumn{2}{c|}{Length 200} & \multicolumn{2}{c}{Length 300}\\
\cmidrule{2-3} \cmidrule{4-5} \cmidrule{6-7}  
& \multicolumn{1}{c}{scTM} & \multicolumn{1}{c|}{scRMSD}
& \multicolumn{1}{c}{scTM} & \multicolumn{1}{c|}{scRMSD}
& \multicolumn{1}{c}{scTM} & \multicolumn{1}{c}{scRMSD}
\\
\midrule
\gthl NativePDBs        &  \gthl 0.91 & \gthl 2.98 & \gthl  0.88 & \gthl 3.24 & \gthl 0.92 & \gthl 3.94\\
\midrule
ESM3(1.4B)              &   0.72 & 13.80  &  0.63  & 21.18 &  0.59 & 25.5 \\
MultiFlow* &  0.86  & 4.73  &  0.86  & 4.98 & 0.86 & 6.01 \\
ProteinGenerator  &   0.91 & 3.75  &   0.88  & 	6.24 &  0.81 & 9.26\\
ProtPardelle      &   0.56 & 12.90  &   0.64  & 13.67 &  0.69 & 14.91\\
\midrule
\chl \method     & \chl  \textbf{0.96} & \chl \textbf{1.80} & \chl \textbf{0.89} & \chl \textbf{4.25}  & \chl \textbf{0.87} & \chl \textbf{5.96} \\
\bottomrule
\end{tabular}
\end{table}

\subsection{Multi-Chain Protein Related Tasks}

\subsubsection{Folding \& Inverse-Folding}

We also initially examined \method's capabilities in modeling multi-chain proteins through folding and inverse-folding tasks. In these tasks, we used samples missing cluster IDs that were dropped during training as the test set, and we also removed samples exceeding a length of 512. The final test set comprised 273 proteins with a number of chains of 2-6. Furthermore, in the two tasks, we only compared \method with two specialized models, Boltz-1~\citep{Boltz-1} and ProteinMPNN, as there are almost no other models that support multi-chain proteins. 
For the inverse-folding task, we employed Boltz-1 with MSA to refold the predicted sequences for calculating \metric{scTM}. As depicted in \cref{tab:multi_chain_folding}, folding for multi-chain proteins represents an extreme challenge. Even with the use of MSA, Boltz-1 exhibits a decline in prediction accuracy compared to single-chain proteins.
Without MSA, achieving effective prediction becomes considerably more difficult.
Although the performance of \method also degrades, it still surpasses that of Boltz-1 without MSA.
Conversely, \method exhibits commendable performance in inverse-folding for multi-chain proteins, with the \metric{scTM} nearly matching the folding performance of Boltz-1 when using reference sequences.

\begin{table}[tb]
    \centering
    \fontsize{9}{9}
    \tabcolsep 3.5pt
    \renewcommand{\arraystretch}{1.0}
    \caption{Performance comparison in multi-chain protein folding (blue highlighted) and inverse-folding tasks (pink highlighted).}
    \label{tab:multi_chain_folding}
   \resizebox{1.0\linewidth}{!}{%
    \begin{tabular}{l|cccc}
\toprule
Method &  \cellcolor{blue_} RMSD $\downarrow$ &  \cellcolor{blue_} TM $\uparrow$ &  \cellcolor{pink_} scTM $\uparrow$ &  \cellcolor{pink_} AAR(\%) $\uparrow$ \\
\midrule
Boltz-1 w/MSA  &     5.40/1.95     &        0.87/0.97   &      -     &     -       \\
Boltz-1 w/oMSA    &     17.86/18.43      &        0.44/0.45   &   -        &     -       \\
ProteinMPNN      &    -       &     -      &   0.90/0.96        &       46.17/46.37     \\
\midrule
\chl \method &   \chl  12.6/13.67     &  \chl   0.64/0.62     &  \chl     0.85/0.95   &   \chl  61.26/59.48    \\
\bottomrule
\end{tabular}
    }
\end{table}


\subsubsection{Multi-Chain Protein Generation}

The most significant difference between the \method and previous methods is its ability to directly generate multi-chain protein complexes. These generated complexes do not require the starting amino acids of each chain to be spatially close to the previous chain. Instead, they have independent spatial positions, yet each chain possesses precise and complementary binding interfaces with the others. However, evaluating these generated complexes poses challenges.
Calculating the self-consistency of the complexes does not accurately assess \method’s capabilities in complex generation as folding models cannot reliably predict structures. Evaluating each single-chain independently for self-consistency is also not entirely appropriate because single-chain proteins may undergo significant conformational changes upon binding, such as 1AKE to 4AKE, or some might only fold correctly in the presence of other proteins.

As \method has demonstrated its single-chain protein generation capability in \cref{exp:single_chain}. We focus on the binding affinity between each chain in multi-chain proteins in this task.
We use $\Delta\text{G}$ to represent the binding strength between two chains of multi-chain proteins with lengths of 50-100, 100-100, and 100-200 (the generation of more chain combinations is shown in the \cref{appendix:experimental_details:multi_chain_lc}).
We report two types of $\Delta\text{G}$: $\Delta\text{G}_\text{RSC}$, for sidechain-only relaxed complexes; and $\Delta\text{G}_\text{RAA}$, for all-atom relaxed complexes (both relaxation and $\Delta\text{G}$ calculation are performed by pyRosetta~\citep{Rosetta,PyRosetta}).
Additionally, we report the \metric{RMSD} between the two structures.
The average/median results are shown in \cref{tab:multi_chain_uncond}. 
For comparison, we utilized Chroma~\cite{Chroma2023} to sample unconditional complex with the same length combinations. 
Besides, we also performed the same task with only \sbm activated ($\method_\text{BB}$), which means generated multi-chain proteins at residue-level.
As shown in \cref{tab:multi_chain_uncond}, compared with using all-atom information, $\method_\text{BB}$ achieved weaker binding strength and higher \metric{RMSD}, which proves \textbf{the importance of the all-atom information in the inter-chain interactions modeling}. 

\begin{table}
    \centering
    \fontsize{9}{9}\selectfont
    \tabcolsep 3.5pt
    \renewcommand{\arraystretch}{1.0}
    \caption{The inter-chain binding affinity between generated complexes. For each metric, we report the average/median value. $\method_\text{BB}$ means using \method in a residue-level manner by only activating the \sbm. We additionally use ProteinMPNN to redesign sequences for Chroma.(marked with *)}
    \label{tab:multi_chain_uncond}
   \resizebox{1.0\linewidth}{!}{%

\begin{tabular}{cllll}
\toprule
\multicolumn{1}{l}{Length} & Model & $\Delta\text{G}_\text{RSC}$ & $\Delta\text{G}_\text{RAA}$  & RMSD     \\
\midrule
\multirow{2}{*}{50-100}             & Chroma             & 133.64/46.51   & -83.96/-86.66   & 1.33/1.22 \\
                                    & Chroma*            & -27.53/-41.71   & -78.41/-77.09   & 1.44/1.28 \\
                                    & \method            & -72.44/-71/91  & -112.65/-116.98 & 1.05/0.95 \\
                                    & \chl $\method_\text{BB}$        & \chl -64.30/-67.30  & \chl -114.94/-114.45 & \chl 1.06/1.03 \\
\midrule
\multirow{2}{*}{100-100}            & Chroma              & 89.47/22.97   & 102.33/48.34   & 1.46/1.39 \\
                                    & Chroma*             & -31.09/-31.51   & -62.15/-59.36   & 1.40/1.30 \\
                                    & \method             & -91.61/-94.54  & -130.31-134.57  & 1.04/0.94 \\
                                    & \chl $\method_\text{BB}$        & \chl -36.74/-69.30  & \chl -117.53/-118.13 & \chl 1.17/1.12 \\
\midrule
\multirow{2}{*}{100-200}            & Chroma              & 79.97/35.86   & -59.32/-54.30   & 1.58/1.48 \\
                                    & Chroma*             & -32.14/-31.79   & -62.79/-59.74   & 1.58/1.39 \\            
                                    & \method             & -44.02/-39.42  & -93.21/-73.09   & 1.35/1.21 \\
                                    & \chl $\method_\text{BB}$          & \chl -3.42/-33.71   & \chl -85.79/-69.12   & \chl 1.58/1.42\\
\bottomrule
\end{tabular}
    }
\end{table}

\begin{figure}[t]
\begin{center}
\centerline{\includegraphics[width=\columnwidth]{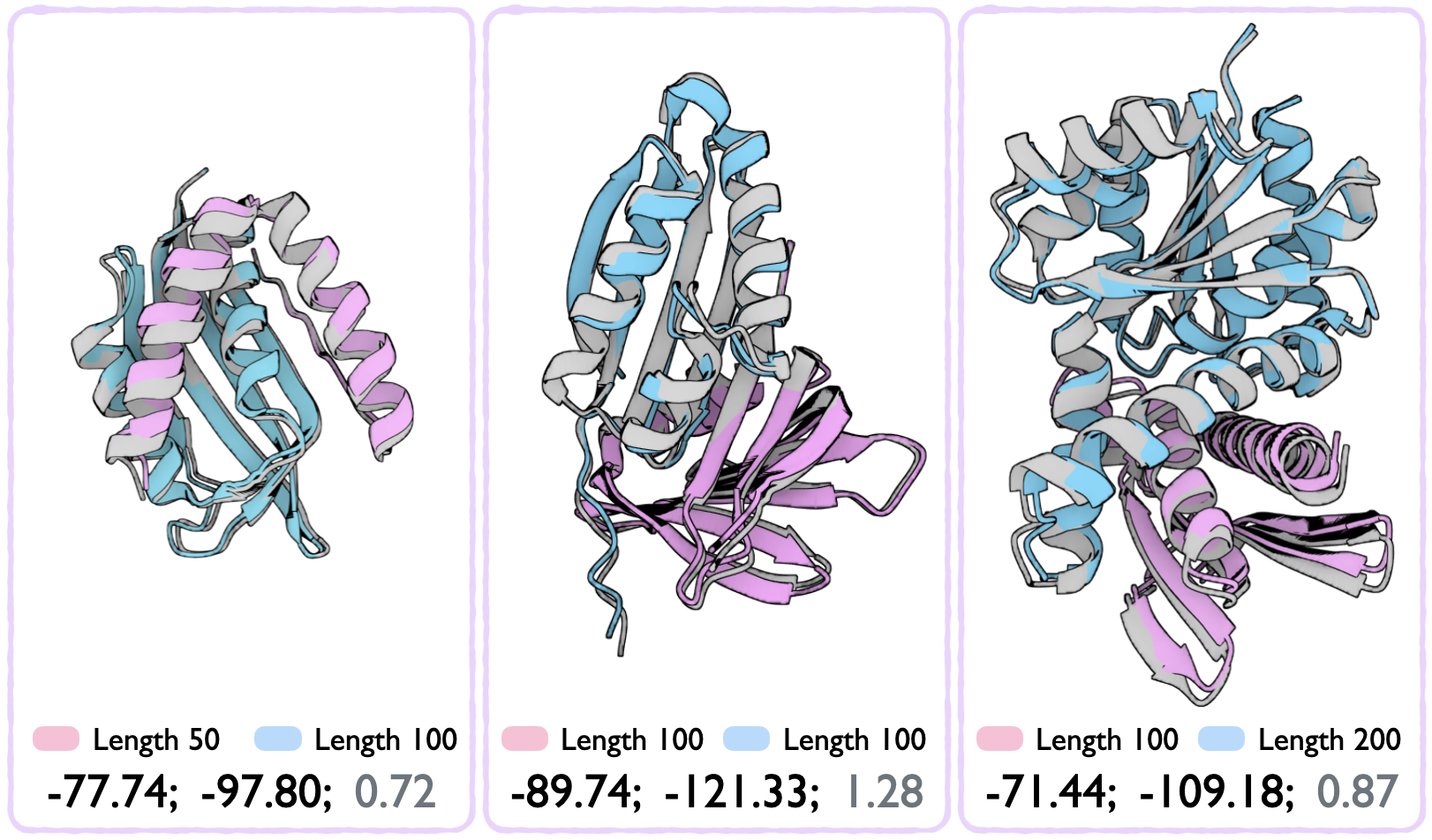}}

\caption{Showcases of the three length combinations. For each case, the gray structure represents \method's generated structure and the colored structure represents the backbone relaxed structure. Different chain is highlighted with different colors. We also report the two $\Delta\textbf{G}$ and the \metric{RMSD} between the two structures.}
\label{fig:showcase}
\end{center}
\end{figure}

By default, \method generates all chains simultaneously. Under this manner, we observed that for complexes with chain length combinations of 50-100 and 100-100, \method tends to generate multi-chain proteins in a ``single-chain protein mode",
leading to strong inter-chain interactions. Conversely, in complexes with chain lengths of 100-200, \method generates two relatively independent single-chain proteins that are bound tightly, resulting in normal binding energies (\cref{fig:showcase}). 
\method also supports an alternative generation manner called \textbf{chain-by-chain}, raised from the conditional generation task in \method's training. The \textbf{chain-by-chain} approach yields significantly different results, with each chain appearing to fold independently before ultimately binding together. Related results can be found in \cref{appendix:experimental_details:multi_chain_cbc}.




\renewcommand{\arraystretch}{0.9}
\begin{table*}[htb]
    \caption{Comprehensive evaluation of peptide design methods across three key aspects: Functionality, Foldability, and Accuracy. The best results are highlighted in \textbf{bold}. 5 out of 93 ground truth samples exhibit $\Delta\text{G}$ greater than 0, are visualized in \cref{appendix:experimental_details:peptide_evaluation}.
    }
    \label{tab:peptide}
    \centering
    \footnotesize
    \begin{tabular}{lrrrrrrr}
\toprule
\multirow{2}{*}{Method} & \multicolumn{2}{c}{Functionality} & \multicolumn{3}{c}{Foldability} & \multicolumn{2}{c}{Accuracy} \\
\cmidrule(lr){2-3} \cmidrule(lr){4-6} \cmidrule(lr){7-8}
& $\Delta\text{G}$ $\downarrow$ & \%~$<$0 $\uparrow$ & pLDDT $\uparrow$ & ipTM $\uparrow$ & Success $\uparrow$ & DockQ $\uparrow$ & \%~$\geq$0.8 $\uparrow$ \\
\midrule
\gthl GroundTruth & \gthl -24.54 & \gthl \textit{94.62} & \gthl 88.31 & \gthl 0.94 & \gthl 100.00\% & \gthl 1.00 & \gthl 100.00 \\
\midrule
PPFlow & -8.56 & 16.72 & 55.72 & 0.57 & 13.01\% & 0.27 & 0.00 \\
DiffPP & -12.40 & 38.17 & 55.10 & 0.57 & 16.55\% & 0.33 & 0.90 \\
PepGLAD & -12.45 & 37.10 & 51.69 & 0.57 & 12.50\% & 0.35 & 0.00 \\
RFDiffusion & -23.27 & \textbf{78.58} & \textbf{69.65} & \textbf{0.73} & \textbf{46.28\%} & 0.28 & 0.00 \\
\midrule
\chl $\method_\text{SFT}$ & \chl -19.90 & \chl 69.34 & \chl 60.36 & \chl 0.66 & \chl 29.22\% & \chl \textbf{0.40} & \chl \textbf{11.29} \\
\chl $\method_\text{zero-shot}$ & \chl \textbf{-23.71} & \chl 62.18 & \chl 60.97 & \chl 0.62 & \chl 27.20\% & \chl 0.24 & \chl 0.12 \\
\bottomrule
\end{tabular}
\end{table*}

\renewcommand{\arraystretch}{1.0}
\begin{table}[t]
    \tabcolsep 7pt
    \centering
    \footnotesize 
    \caption{Performance comparison of antibody design methods on RAbD benchmark. The best results are shown in \textbf{bold}.
    }
    \resizebox{1.0\linewidth}{!}{%
    \begin{tabular}{lrrrr}
\toprule
Method & AAR (\%) $\uparrow$ & RMSD $\downarrow$ & $\text{E} \downarrow$ & $\Delta\text{G} \downarrow$ \\
\midrule
\gthl RAbD & \gthl 100.00 & \gthl 0.00 & \gthl -16.76 & \gthl -15.33 \\
\midrule
dyMEAN & 40.05 & 2.36 & 1239.29 & 612.75 \\
DiffAb & 35.04 & 2.53 & 495.69 & 489.42 \\
AbDPO & 31.29 & 2.79 & 270.12 & 116.06 \\
AbDPO++ & 36.25 & 2.48 & 338.14 & 223.73 \\
\midrule
\chl $\method_\text{SFT}$ & \chl \textbf{41.20} & \chl \textbf{2.08} & \chl \textbf{137.74} & \chl 91.64 \\
\chl $\method_\text{zero-shot}$ & \chl 28.35 & \chl 5.81 & \chl 284.24 & \chl \textbf{81.12} \\
\bottomrule
\end{tabular}
    }
    \label{tab:antibody}
\end{table}

\subsection{Downstream Tasks}

We further verify the capacity of \method on specific tasks including antibody design and peptide design,
in both supervised fine-tuning (SFT) and zero-shot manner.
Details about SFT refer to \cref{appendix:training_details:training,appendix:experimental_details:antibody,appendix:experimental_details:peptide}.


\subsubsection{Antibody CDR-H3 Co-Design}

\paragraph{Setup.}
The design of Complementarity Determining Regions(CDRs) is a crucial step in developing potent therapeutic antibodies, especially CDR-H3. 
Following the data preprocessing pipeline introduced in~\citep{ProteinBench,AbDPO}, we conduct training on the Structural Antibody Database~\citep{SAbDab} and perform evaluation on the RAbD benchmark~\citep{RAbD}.

We compare our model with four antigen-specific antibody design methods (\textbf{dyMEAN}~\citep{dyMEAN}, \textbf{DiffAb}~\citep{DiffAb}, \textbf{AbDPO}~\citep{AbDPO} and its variant \textbf{AbDPO++}). 
Following previous works~\citep{ProteinBench}, we use multiple metrics to evaluate the quality of designed CDRs: \metric{AAR} and C$_\alpha$ \metric{RMSD} for generated sequence and backbone structure; 
\metric{Total Energy ($\text{E}$)} and \metric{Binding Energy} ($\Delta \text{G}$) for atomic rationality and functionality, which are provided by pyRosetta.


\paragraph{Results.}
As indicated in \cref{tab:antibody}, \method performs significantly superior to other methods in all metrics.
With respect to accuracy related metrics, including \metric{AAR} and \metric{RMSD}, \method's superior performance demonstrates its capability in generating antibodies that resemble natives.
On rational and functional metrics, \method's generated antibodies exhibit the highest rationality and binding capability.
Additionally, antibodies generated in a zero-shot manner display excellent $\Delta\text{G}$, proving \method's capacity in inter-chain interacted protein generation, while the abnormal \metric{AAR}/\metric{RMSD} highlight the different binding patterns between general proteins and antibodies (refer to \cref{appendix:experimental_details:antibody_zero-shot} for details).

\subsubsection{Peptide Design}
\paragraph{Setup.} 
The design of functional and binding peptides plays a crucial role in pharmacological applications and targeted therapeutic development. 
To evaluate our \method's performance in receptor-targeted peptide design, we use the \textbf{PepBench}~\citep{PepGLAD} dataset for training and validation and use the \textbf{LNR}~\citep{LNR} dataset as the test set.

We compare our model with: \textbf{PepGLAD}~\citep{PepGLAD}, \textbf{PPFlow}~\citep{PPFlow}, and \textbf{DiffPP}~\citep{PPFlow}.
We also include \textbf{RFDiffusion}~\citep{RFdiffusion}, which utilizes ProteinMPNN for sequence design. 
The peptide candidates are comprehensively evaluated across three key aspects: \textbf{Functionality}, \textbf{Foldability}, and \textbf{Accuracy}.
For functionality, we evaluate the binding energy ($\Delta\text{G}$)
and the proportion of candidates with $\Delta\text{G}$ below zero, \%~$<$0.
For foldability, we use Boltz-1(wMSA) to fold sequences of generated peptides, then evaluate the folded structure 
in two confidence metrics: predicted Local Distance Difference Test (\metric{pLDDT}) and interface predicted Template Modeling (\metric{ipTM})~\citep{zhang2004scoring,xu2010significant} score. 
We also report a comprehensive metric, defined as the proportion of candidates with both a \metric{pLDDT} score $\ge 70$ and an \metric{ipTM} score $\ge 0.8$, donated as Success.
For accuracy, we evaluate the \metric{DockQ}~\citep{DockQ,DockQv2} score and the
proportion of candidates achieving a \metric{DockQ} score of at least 0.8 (\%~$\geq$0.8), which is the threshold considered as high-quality.

\paragraph{Results.}
As indicated in \cref{tab:peptide}, \method exhibits competitive performance across all three key aspects.
 For functionality, \method generates peptides with an average binding energy of -19.90 and achieves negative $\Delta\text{G}$ in 69.34\% samples, significantly outperforming other methods.
The performance in foldability metrics demonstrates \method's superiority in sequences generation, with the high \metric{pLDDT} and \metric{ipTM}. 
For accuracy, \method stands out as nearly the only method capable of generating peptides with a \metric{DockQ} score exceeding 0.8.
Additionally, the peptides generated by \method in a zero-shot manner perform similarly to antibodies, with high affinity but do not resemble natural ones. This is reflected in the comparable performance in functionality \& foldability and the significant degradation in accuracy. 
We also present more results on longer binder design in ~\cref{appendix:binder_design}.

\section{Discussions}
\label{sec:conclusion}
In this paper, we introduce \method, a generative model for protein complexes designing at all-atom level.
\method is capable of generating tightly bound protein complexes, executing high-quality single-chain protein-related tasks, and achieving remarkable performance in specific functional protein design tasks.
Despite \method's potential in AI-based functional protein design tasks, several limitations remain to be addressed.
These limitations are primarily in these aspects:
(1) the performance in folding tasks requires further improvement;
(2) the functionality of \rfm is relatively restricted;
(3) the number of downstream tasks is limited.
Future works are detailed in the \cref{appendix:limitations}.


\section*{Impact Statement}
Our work on multi-chain protein generation can be used in developing potent therapeutic macromolecues such as antibodies and accelerating the research process of drug discovery. 
Our method may be adapted to other scenarios of computer-aided design, such as small molecule design, material design, and chip design. 
It is also needed to ensure the responsible use of our method and refrain from using it for harmful purposes.

\section*{Acknowledgements}
We thank anonymous reviewers for their insightful feedback. 
We would like to especially thank Dr. Hang Li for insightful discussions on the project and feedback on the manuscript that help shape this study. 
We thank Yi Zhou, Yuning Shen, Xinyou Wang, Yiming Ma, Fei Ye, and Lihao Wang for their valuable comments.

\bibliography{references}
\bibliographystyle{icml2025}

\newpage
\clearpage
\appendix
\onecolumn
\section{Model}
\label{appendix:model}

\subsection{$\text{SE}(3)$ Flow Matching}
\label{appendix:model:se3flow}

Flow Matching (FM)~\citep{FlowMatching} offers an efficient framework for learning continuous normalizing flows by directly learning a time-dependent vector field that transforms samples from a prior distribution to a target data distribution, eliminating the need for expensive likelihood evaluations or ODE solving during training.

Consider a prior distribution $p_0$ and a target distribution $p_1$. 
FM learns a time-dependent vector field $\mathbf{v}_t: \mathbb{R}^d \times [0,1] \rightarrow \mathbb{R}^d$ that guides the transformation through a continuous-time flow $\psi_t: \mathbb{R}^d \rightarrow \mathbb{R}^d$, governed by the ordinary differential equation (ODE):

\begin{equation}
\frac{d}{dt}\psi_t(\mathbf{x}) = \mathbf{v}_t(\psi_t(\mathbf{x})), \quad \psi_0(\mathbf{x}) = \mathbf{x}
\end{equation}

where samples $\mathbf{x} \sim p_0$ are drawn from the prior distribution and transformed to follow the target distribution $p_1$ at $t=1$.

To enable tractable training, FM introduces an \textit{interpolant} $\phi_t(\mathbf{x}_0, \mathbf{x}_1)$ that defines a smooth path between pairs of points $\mathbf{x}_0 \sim p_0$ and $\mathbf{x}_1 \sim p_1$. The conditional vector field $\mathbf{u}_t$ is derived as the time derivative of this interpolant. 
The conditional flow matching objective then becomes:

\begin{equation}
\mathcal{L}_{\text{CFM}}(\theta) = \mathbb{E}_{t\sim\mathcal{U}[0,1], \mathbf{x}_0\sim p_0, \mathbf{x}_1\sim p_1}\left[\|\mathbf{v}_\theta(\mathbf{x}_t, t) - \mathbf{u}_t(\mathbf{x}_t|\mathbf{x}_0,\mathbf{x}_1)\|^2\right]
\end{equation}

where $\mathbf{x}_t = \phi_t(\mathbf{x}_0, \mathbf{x}_1)$. After training, new samples are generated by solving:

\begin{equation}
\frac{d}{dt}\mathbf{x}_t = \mathbf{v}_\theta(\mathbf{x}_t, t), \quad \mathbf{x}_0 \sim p_0
\end{equation}

When applying this framework to $\text{SE}(3)$, we need to consider both translations and rotations. 
For translations in $\mathbb{R}^3$, we employ linear interpolation:

\begin{equation}
\phi_t^{\text{trans}}(\mathbf{x}_0^i, \mathbf{x}_1^i) = (1-t)\mathbf{x}_0^i + t\mathbf{x}_1^i, \quad \mathbf{x}_0^i, \mathbf{x}_1^i \in \mathbb{R}^3
\end{equation}

with the corresponding conditional vector field:

\begin{equation}
\mathbf{u}_t^{\text{trans}}(\mathbf{x}_t^i|\mathbf{x}_0^i,\mathbf{x}_1^i) = \mathbf{x}_1^i - \mathbf{x}_0^i = \frac{\mathbf{x}_1^i - \mathbf{x}_t^i}{1-t}
\end{equation}

For rotations in $\text{SO}(3)$, following~\citep{Riemannianflow,Multiflow,FrameFlow}, we utilize geodesic interpolation on the manifold. During training, we use a linear schedule:

\begin{equation}
\phi_t^{\text{rot}}(R_0^i, R_1^i) = \text{exp}_{R_0^i}(t \cdot \text{log}_{R_0^i}(R_1^i)), \quad R_0^i, R_1^i \in \text{SO}(3)
\end{equation}

where $\exp (\cdot)$ and $\log (\cdot)$ denote the exponential and logarithm maps on $\text{SO}(3)$. 
During inference, we employ an exponential schedule $\kappa(t) = e^{-ct}$ with $c=10$:

\begin{equation}
\phi_t^{\text{rot}}(R_0^i, R_1^i) = \text{exp}_{R_0^i}((1-e^{-ct})\text{log}_{R_0^i}(R_1^i))
\end{equation}

The conditional vector field in the tangent space $T_{R_t}\text{SO}(3)$ takes different forms during training and inference. During training, following the linear schedule, it is given by:

\begin{equation} 
\mathbf{u}_t^{\text{rot}}(R_t^i|R_0^i,R_1^i) = \frac{\text{log}_{R_t^i}(R_1^i)}{1-t}
\end{equation}

while during inference, with the exponential schedule, it becomes:

\begin{equation} 
\mathbf{u}_t^{\text{rot}}(R_t^i|R_0^i,R_1^i) = c\text{log}_{R_t^i}(R_1^i)
\end{equation}

For the choice of distributions, we consider the geometric properties of $\text{SO}(3)$. 
We use the uniform distribution over $\text{SO}(3)$ during training and sampling.

For training, we define separate loss terms that jointly guide the learning of the vector field. The translation loss follows the Euclidean Flow Matching objective:

\begin{equation}
\mathcal{L}_{\text{trans}}(\theta) = \mathbb{E}_{t\sim\mathcal{U}[0,1],\mathbf{x}_0,\mathbf{x}_1}\left[\frac{1}{N} \sum_{i=1}^{N} \|\mathbf{v}_\theta^{\text{trans}}(\mathbf{x}_t^i, t) - \mathbf{u}_t^{\text{trans}}(\mathbf{x}_t^i|\mathbf{x}_0^i,\mathbf{x}_1^i)\|^2\right]
\end{equation}

For rotations, following the Riemannian geometry of $\text{SO}(3)$, we define the loss using the geodesic distance on the Lie algebra:

\begin{equation}
\mathcal{L}_{\text{rot}}(\theta) = \mathbb{E}_{t\sim\mathcal{U}[0,1],R_0,R_1}\left[\frac{1}{N} \sum_{i=1}^{N} \left\|\mathbf{v}_\theta^{\text{rot}}(R_t^i, t) - \mathbf{u}_t^{\text{rot}}(R_t^i|R_0^i,R_1^i)\right\|^2_{\text{SO}(3)}\right]
\end{equation}

where $R_t^i = \phi_t^{\text{rot}}(R_0^i, R_1^i)$ represents the interpolated rotation at time $t$, and $\mathbf{v}_\theta^{\text{rot}}(R_t^i, t) \in \mathfrak{so}(3)$ is the predicted velocity in the Lie algebra. The complete $\text{SE}(3)$ Flow Matching objective combines both terms:

\begin{equation}
\mathcal{L}_{\text{SE}(3)}(\theta) = \mathcal{L}_{\text{trans}}(\theta) + \mathcal{L}_{\text{rot}}(\theta)
\end{equation}

During inference, we solve the ODE using the exponential schedule for rotations while maintaining the linear schedule for translations.

\subsection{Discrete Flow Matching}
\label{appendix:model:discreteflow}



While continuous Flow Matching effectively handles continuous data in $\mathbb{R}^3$ and $\text{SO}(3)$, discrete data such as amino acid sequences require a different approach. 
We adopt Discrete Flow Matching~\citep{Multiflow}, and define a path from a masked token distribution to the data distribution. 
Let $\bm{S}_1 = [S_1^1, ..., S_1^N]$ be a sequence from the data distribution, and $M$ denote the mask token. 
The interpolant between $\bm{S}_1$ and the fully masked sequence is a categorical distribution defined via the Kronecker delta. 
We define the $i$-th token $\bm{S}_t^i$ at intermediate time $t$ as:
\begin{equation}
p_{t|1}(\bm{S}_t^i|\bm{S}_1^i) = t\delta\{\bm{S}_t^i, \bm{S}_1^i\} + (1-t)\delta\{\bm{S}_t^i, M\}
\end{equation}
where $\delta\{a, b\} = 1$ if $a = b$ and $0$ otherwise. This interpolant linearly mixes the clean sequence and the mask state over time.

The loss function is the cross-entropy between the predicted and data distributions, written as:
\begin{equation}
\mathcal{L}_{\text{discrete}}(\theta) = \mathbb{E}_{\substack{t \sim \mathcal{U}[0,1] \\ \bm{S}_t \sim p_{t|1}(\cdot|\bm{S}_1)}} \left[  -\log p_{1|t}^\theta(\bm{S}_1|\bm{S}_t) \right]
\end{equation}
where $\bm{S}_t \sim p_{t|1}(\cdot|\bm{S}_1)$ samples a corrupted sequence at time $t$ using the above conditional interpolant, $p_{1|t}^\theta(\bm{S}_1|\bm{S}_t)$ is the neural network's predicted distribution given the corrupted sequence $\bm{S}_t$.

\subsection{Sub-module Architectures}
\label{appendix:submodule_structure}
The detailed structure is shown in \cref{fig:illustration}. In \scm, the protein language model is only activated in training phase II, and the protein language model encoding is weighted with a learnable zero-initialized parameter before merging into residue representation. 

The resdiue level and pair-residue level information are encoded in 384 and 192 dim in \sbm and \rfm, while the dims are 256 and 128 in \scm. For model size, the overall \method contains 127M parameters, of which \sbm contains 52M, \scm contains 22M, and \rfm contains 54M parameters.

\rfm is initialized to apply zero change to the \sbm generated protein to ensure that \rfm does not undergo negative optimization.

\section{Training Details}
\label{appendix:training_details}

\subsection{Loss for \sbm}
\label{appendix:training_details:bb_model}

\paragraph{Flow-matching Loss.}
The detailed flow-matching loss, $\mathcal{L}_\text{flow-matching}$, refers to \cref{appendix:model:se3flow} and \cref{appendix:model:discreteflow}.

\paragraph{Consistency Loss.}
For arbitrary $t$ for each modality ($t_S$ for sequence, $t_T$ for backbone structure), we can get the corresponding noised data ($\bm{S}_{t_S}$ means noised sequence with $t_S$, $\bm{T}_{t_T}$ means noised structure with $t_T$) and predict the final sample by the \sbm in \method, $\method_\text{Seq\&BB}$:
\begin{equation}
\hat{L}_1, \hat{\bm{T}}_1 = \text{\method}_\text{Seq\&BB}(\bm{S}_{t_S}, \bm{T}_{t_T}, t_S, t_T)
\end{equation}
We can also get the noised data from the adjacent $t$, $S_{t_S+\Delta t}$ and $T_{t_T+\Delta t}$, and predict the final sample:
\begin{equation}
\hat{L}_1', \hat{\bm{T}}_1' = \text{\method}_\text{Seq\&BB}(\bm{S}_{t_S+\Delta t}, \bm{T}_{t_T+\Delta t}, t_S+\Delta t, t_T+\Delta t)
\end{equation}
Consistency loss is defined as the gap between the two predictions on the two modalities. Since the predictions from the $t$ closer to 1 are more accurate, we set the predictions from $t+\Delta t$ as a teacher. For sequence, the gap is the KL divergence between the two predicted logits:
\begin{equation}
\mathcal{L}_{\text{consistency\_S}} = \text{KL}\Bigl(\log\bigl(\text{softmax}(\hat{L}_1)\bigr), \log\bigl(\text{softmax}(\hat{L}_1^{\prime})\text{.detach()}\bigr)\Bigr)
\end{equation}
For structure, the gap is the MSE between the two predicted structures:
\begin{equation}
\mathcal{L}_{\text{consistency\_T}} = \text{MSE}\Bigl(\text{trans}(\hat{\bm{T}}_1), \text{trans}(\hat{\bm{T}}_1^{\prime})\text{.detach()}\Bigr) + \text{MSE}\Bigl(\text{Mat2Vec}\bigl(\text{rot}(\hat{\bm{T}}_1)\bigr), \text{Mat2Vec}\bigl(\text{rot}(\hat{\bm{T}}_1^{\prime})\bigr)\text{.detach()}\Bigr)
\end{equation}
Considering that the quality of the \method's predictions is not high when $t$ approaches 0, it is unreasonable to demand consistency at this point. Thus, we scale the consistency loss with respect to $t$, reducing the impact of consistency loss on model training when $t$ is small. Additionally, we also followed the construction method for consistency loss proposed by \citet{improvedCT}. Finally, the consistency loss used in \sbm is defined as:
\begin{equation}
\mathcal{L}_\text{consistency}=t^2_S\times(\sqrt[2]{{\mathcal{L}_{\text{consistency}\_S}}^2+c_S^2}-c_S)+t^2_T\times(\sqrt[2]{{\mathcal{L}_{\text{consistency}\_T}}^2+c_T^2}-c_T)
\end{equation}
\begin{equation}
c_S=0.00054 * \sqrt[2]{\text{dim}_S}, \quad c_T=0.00054 * \sqrt[2]{\text{dim}_T}
\end{equation}

\subsection{Loss for \scm}
\label{appendix:training_details:sc_model}
We followed AlphaFold2~\citep{AlphaFold2} to build the loss for \scm, $\mathcal{L}_\chi$ and $\mathcal{L}_\text{FAPE}$. $\mathcal{L}_\chi$ is built according to \metric{Algorithm 27} in the supplementary information of AlphaFold2 and $\mathcal{L}_\text{FAPE}$ is built according to \metric{Algorithm 28}.


\subsection{Loss for \rfm}
\label{appendix:training_details:rf_model}

\paragraph{Correction Loss.}
The correction loss, $\mathcal{L}_\text{corr}$, is similar to \sbm's flow-matching loss.

\paragraph{Auxiliary Loss.}
The auxiliary loss consists of backbone FAPE loss, $\mathcal{L}_\text{BB-FAPE}$, and residue distogram prediction loss, $\mathcal{L}_\text{dist}$.

$\mathcal{L}_\text{BB-FAPE}$ is the simplified version of $\mathcal{L}_\text{FAPE}$ which only considers the backbone atoms.

$\mathcal{L}_\text{dist}$ is the gap between the real residue-pair distance and the generated residue-pair distance. For a protein with the length of $N$, the distance between any two residues $i$ and $j$ is donated as $d_{ij}$. The $\mathcal{L}_\text{dist}$ is defined as:
\begin{equation}
\mathcal{L}_\text{dist} = \frac{1}{N(N-1)}\sum_{i=1}^{N}\sum_{j=1,j\ne i}^{N}||d^{ij}-\hat{d}^{ij}||^2
\end{equation}

\subsection{Training}
\label{appendix:training_details:training}
\paragraph{training phase I.}
In training phase I, the \sbm was trained on 64$\times$H100 GPUs with 257,000 steps,
with a learning rate of 1e-4. The \scm was trained on 8$\times$H100 GPUs, accumulating a total of 836,901 steps, also with a learning rate of 1e-4. 

\paragraph{training phase II.}
In training phase II, \method was trained on 64$\times$H100 GPUs with 235,000 steps
. The learning rate for the \sbm is set to 1e-5. 

\paragraph{SFT.}
In the SFT phase, we used 8$\times$H100 GPUs to fine-tune \method for antibody design and peptide design. The SFT phase lasted for 1200 epochs for every task. The learning rate for each module is set to 5e-5 and the training cycle is adjusted to 10-1-1.

\paragraph{Protein Language Model.} We used a drop-in replacement for the ESM protein language model implementation named FAESM~\cite{FAESM}. We thank the authors for providing an efficient FlashAttention-based~\cite{dao2022flashattention} implementation, which significantly accelerated the training speed.

\section{Sidechain Torsion Angles Distribution}
\label{appendix:tor_distribution}

\begin{figure*}
\begin{center}
\centerline{\includegraphics[width=0.9\textwidth]{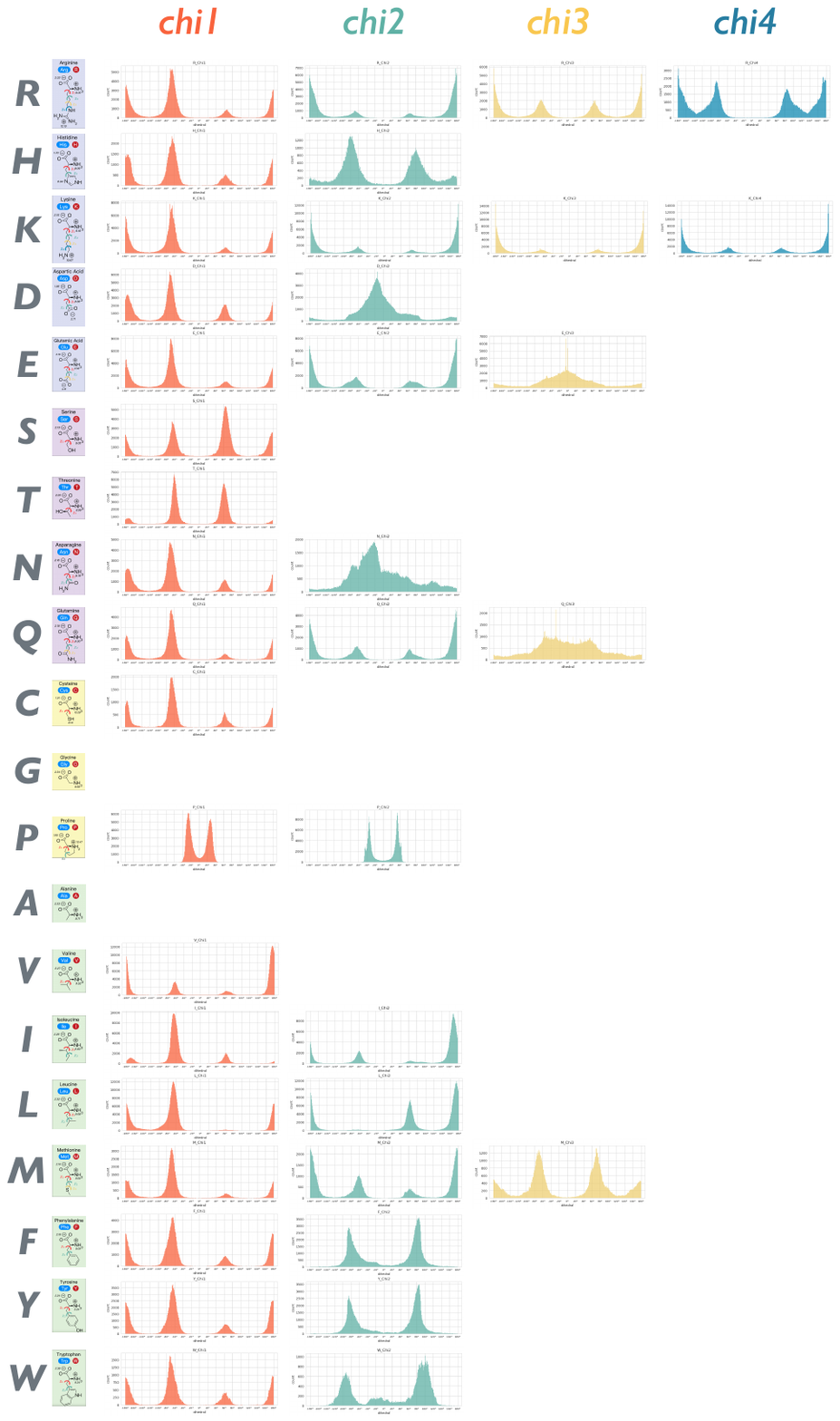}}
\vspace{-2mm}
\caption{The distribution of four sidechain torsion angles in all amino acid types.}
\label{fig:chi_dist}
\vspace{-8mm}
\end{center}
\end{figure*}

We calculated the sidechain torsion angles for each amino acid in 22,281 single-chain proteins sourced from the PDB, with protein lengths ranging from 50 to 20,000.
Subsequently, we analyzed the distribution of each sidechain torsion angle for each type of amino acid. 
As shown in \cref{fig:chi_dist}, it is evident that the number and distribution of sidechain torsion angles differ among various types of amino acids. 
There are exceptions, such as phenylalanine (\textbf{F}), tyrosine (\textbf{Y}), and tryptophan (\textbf{W}), which have similar sidechain torsion angle distributions due to their structural similarity. In the BLOSUM62 matrix \citep{BLOSUM}, the substitution scores between these three amino acids are positive, indicating that they can be substituted with each other to some extent. 
Therefore, the sidechain torsion angles retain substantial information about the amino acid types even being noised.

\section{Experimental Details}
\label{appendix:experimental_details}

\subsection{Sequence Sampling}
\label{appendix:experimental_details:seq_sampling}

The predicted sequence $\hat{\bm{S}}_1$ is sampled from the predicted logits $\hat{L}_1$. 
If $\method_\text{Refine}$ is activated, the $\hat{L}_1$ comes from two modules, \sbm and \rfm:
\begin{equation}
\hat{L}_1=
\begin{cases}
\method_\text{Seq\&BB}(\bm{S}_{t_S}, \bm{T}_{t_T}, t_S, t_T), & \text{if } t_S<0.8 \\
0.8\times\method_\text{Seq\&BB}(\bm{S}_{t_S}, \bm{T}_{t_T}, t_S, t_T)+0.2\times\method_\text{Refine}(\hat{\bm{S}}_1, \hat{\bm{T}}_1, t_S,t_T,\hat{\bm{\chi}}), & \text{if } t_S\ge0.8 
\end{cases}
\end{equation}
Then, the $\hat{\bm{S}}_1$ is sampled following:
\begin{equation}
\hat{S}_1=
\begin{cases}
\text{Categorical}(\text{Softmax}(\hat{L}_1/\mathcal{T})), & \text{if } t_S<0.85 \\
\arg \max(\hat{L}_1), & \text{if } t_S\ge0.85
\end{cases}
\end{equation}
where the temperature $\mathcal{T}$ follows an exponential decay schedule:
\begin{equation}
\mathcal{T} = \mathcal{T}_{\max} \times \exp(-\lambda \times t_S)
\end{equation}
with hyperparameters $\mathcal{T}_{\max}=30$ and decay rate $\lambda=30$.
When noising the $\hat{S}_1$ for the next step, we sort all the positions with their scores and only keep the amino acid with the top $K$ scores, for a protein with the length of $N$, $K=\text{int}(t_S\times N)$. The score, $\mathcal{S}^i$, for any position $i$ is defined as :

\begin{equation}
\mathcal{S}^i=\log(\text{Softmax}(\hat{L}^{i}_1))[\hat{\bm{S}}^{i}_1] + (1-t_S)\times\log(\log(\mathcal{R}+10^{-8})+10^{-8}),\quad \mathcal{R}\sim \mathcal{N}(0, 1)
\end{equation}

where $\log(\log(\mathcal{R}+10^{-8})+10^{-8})$ is a random term used to avoid decoding sequences in local optima.
We denote the score of the $K$th highest as $\mathcal{S}_K$, then the sequence for the next sampling step is defined as:

\begin{equation}
\hat{\bm{S}}_{t_S+\Delta t}=\{\hat{\bm{S}}_{t_S+\Delta t}^i|i\in[1,N]\},\qquad \hat{\bm{S}}^{i}_{t_S+\Delta t}=
\begin{cases}
\hat{\bm{S}}^{i}_1,& \text{if } \mathcal{S}^i\ge\mathcal{S}_K \\
\texttt{[MASK]},& \text{if } \mathcal{S}^i<\mathcal{S}_K
\end{cases}
\end{equation}


\subsection{Statistical Validation on Folding}

We conducted folding with APM and ESM3 using 20 random seeds. For RMSD, ESM3 shows a marginally better mean (4.708±0.094 vs 4.828±0.077) with statistical significance (p $<$ 0.05). For TM-score, APM achieves better performance (0.856±0.002 vs 0.828±0.002) with statistical significance (p $<$ 0.05). The detailed results are shown below with the format of (average±std).

\begin{table}[ht]
\centering
\caption{Folding performance comparison between ESM3 and APM}
\label{tab:folding_comparison}
\begin{tabular}{lcc}
\hline
Method & RMSD ~$\downarrow$  & TM~ $\uparrow$ \\
\hline
ESM3 (1.4B) & 4.708±0.094 & 0.828±0.002 \\
APM & 4.828±0.077 & 0.856±0.002 \\
\hline
\end{tabular}
\end{table}

\subsection{Multi-Chain Protein}
\label{appendix:experimental_details:multi_chain}

\subsubsection*{Multi-Chain Protein Generation without All-Atom}
\label{appendix:experimental_details:multi_chain_bb}


During the phase II of \method training, the loss for the \sbm is computed directly based on its own output, rather than relying on the output from the \rfm and then back-propagating to \sbm. As a result, once the entire training process is completed, the \sbm can be used independently, allowing for protein generation at the residue level. 
To verify the importance of all-atom information in multi-chain protein design, we conducted an ablation study by generating multi-chain proteins at the residue level using only the \sbm, and report the results in the main text (\cref{tab:multi_chain_uncond}). Although the backbone structure tends to stabilize when the \scm and \rfm are activated during the last 20\% of steps, the two different versions of the model still exhibit significant energy differences.

When the design is carried out at the residue level, the inter-chain binding strength significantly decreases. Additionally, the difference in $\Delta\text{G}$ and structure before and after all-atom relaxation becomes more pronounced, indicating that information at the amino acid level alone is insufficient for modeling multi-chain interactions, as opposed to the complete \method.

Furthermore, for proteins with lengths of 100-200, without using all-atom information, $\Delta\text{G}$ of the model-generated structures yields a mean value close to 0 and a median of -33. This suggests the presence of clashes at the binding interfaces, a situation not observed in the complete \method. These findings underscore the importance of all-atom information in modeling inter-chain protein interactions.

\subsubsection*{More Chain Length Combinations}
\label{appendix:experimental_details:multi_chain_lc}

\begin{figure*}[t]
\begin{center}
\centerline{\includegraphics[width=1.0\textwidth]{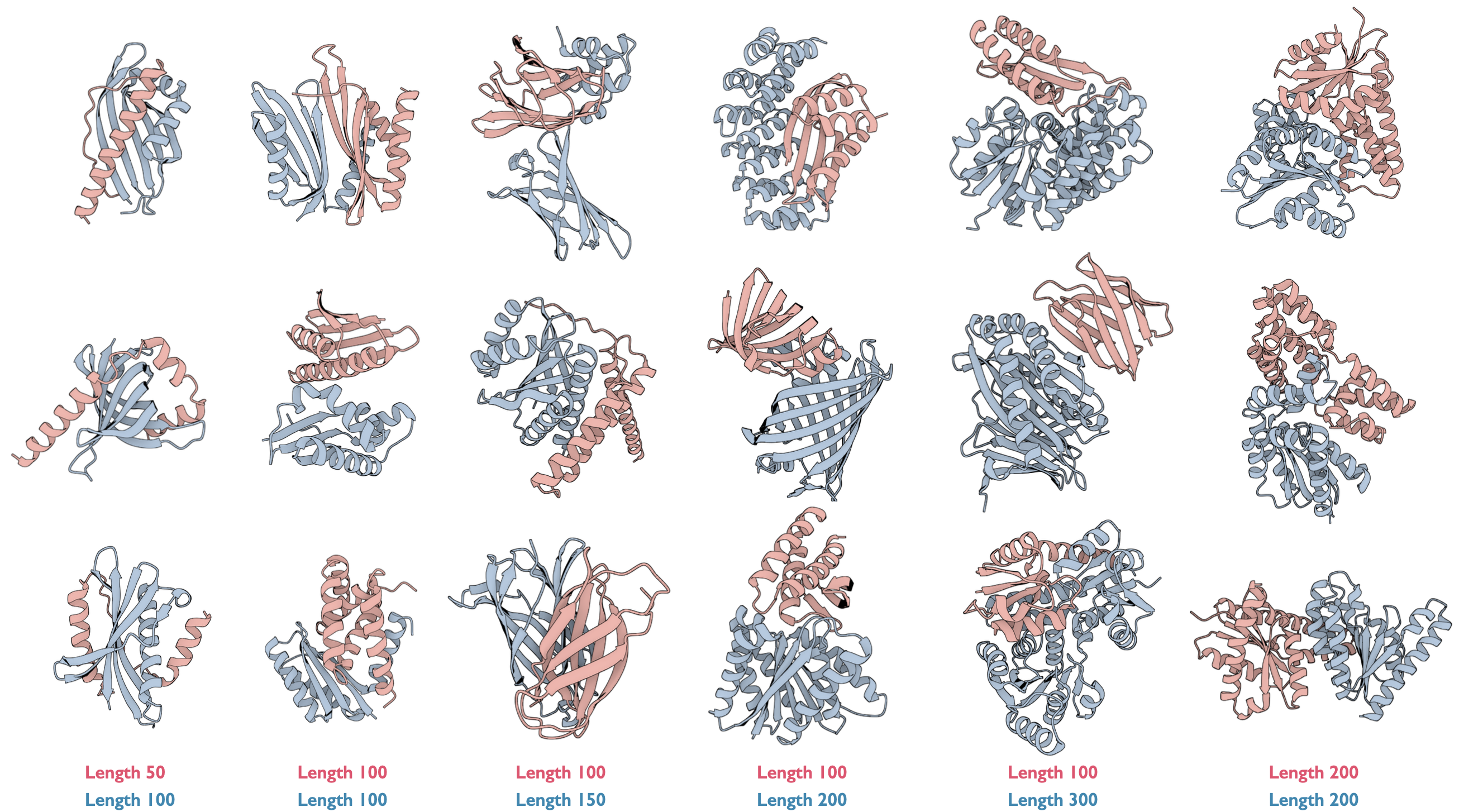}}
\caption{\method generated proteins with chain lengths of 50-100, 100-100, 100-150, 100-200, 100-300, and 200-200.}
\label{fig:all_cases}
\vspace{-5mm}
\end{center}
\end{figure*}

\paragraph{Length Combinations.}
In the main text, we have reported the results of multi-chain protein generation with chain length combinations of 50-100, 100-100, and 100-200. Here, we present metrics for additional chain length combinations, including 50-200, 100-150, 100-300, and 200-200 in \cref{tab:multi_chain_uncond_supp}. Furthermore, we show more cases of each length combination in \cref{fig:all_cases}.

\begin{table}[h]
    \vspace{0mm}
    \centering
    \fontsize{9}{9}\selectfont
    \tabcolsep 10pt
    \renewcommand{\arraystretch}{1.0}
    \caption{$\Delta\text{G}$ and the \metric{RMSD} before and after all-atom relaxation for more chain length combinations. The trend of affinity change is consistent with that in the main text.}
    \label{tab:multi_chain_uncond_supp}
    \begin{tabular}{llll}
\toprule
\gthl Length  & $\Delta\text{G}_\text{RSC}$    & $\Delta\text{G}_\text{RAA}$   & RMSD      \\
\midrule
\gthl 100-150 & -63.34/-53.04 & -102.74/-90.41  & 1.28/1.10 \\
\gthl 50-200  & -42.38/-63.71 & -117.19/-110.37 & 1.19/1.10 \\
\gthl 200-200 & -28.41/-49.67 & -90.10/-83.55   & 1.61/1.37 \\
\gthl 100-300 & -23.43/-28.67 & -73.03/-60.93   & 1.78/1.46 \\
\bottomrule
\end{tabular}
    \vspace{2mm}
\end{table}

\begin{figure*}
\begin{center}
\centerline{\includegraphics[width=1.0\textwidth]{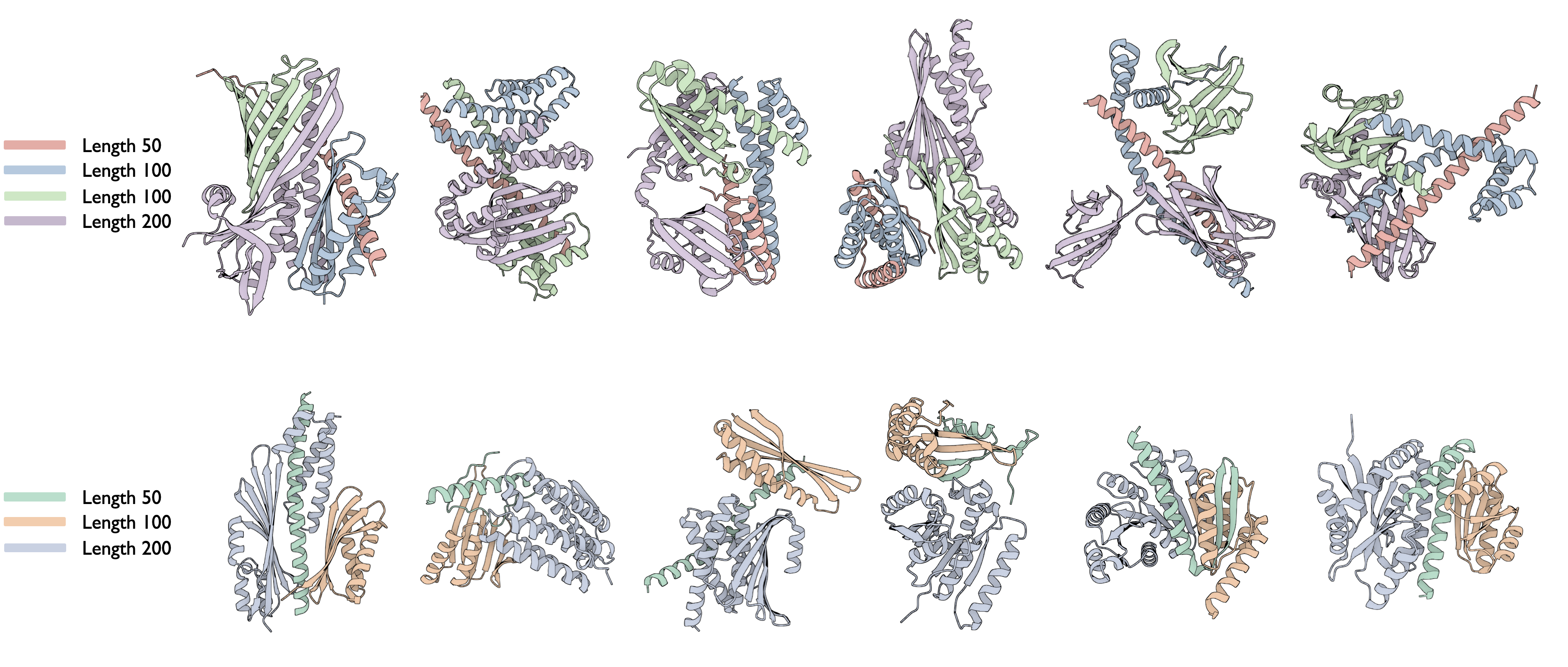}}
\caption{\method generated proteins with more than 2 chains. The top row, generated protein complexes composed of 4 chains; the bottom row, generated protein complexes composed of 3 chains. The length of each chain is highlighted with a unique color.}
\label{fig:3-4_chains}
\end{center}
\end{figure*}

\paragraph{Chain Numbers.}
Theoretically, \method supports the generation of protein complexes composed of any number of chains. However, we observed that \method's performance declines when the generated complex contains more than two chains. 
This is evident in the increased presence of unstructured backbones or abnormally high ratios of $\alpha$-helix/$\beta$-sheet structures (over 90\% of single secondary structure) in the generated complexes.
Interestingly, these structures still exhibit no obvious clashes in the binding interface. We do not apply detailed metric calculations since $\Delta\text{G}$ only computes the binding strength between two components. Instead, we present some cases for proteins composed of more chains in \cref{fig:3-4_chains}.

We attribute the degradation in performance of \method with more chains to two main reasons: \textbf{1)} The majority of the complex samples consist of two chains, and complexes with more chains are significantly less; \textbf{2)} With more chains, the overall length of the protein increases, which also results in reduced model generation quality.

\subsubsection*{Chain-by-Chain Generation}
\label{appendix:experimental_details:multi_chain_cbc}

By default, \method generates all the chains within a multi-chain protein simultaneously. However, we have also implemented an iterative generation manner, called ``\textbf{chain-by-chain}". \method supports this manner because its training task of performing conditional generation, which means generating the remaining chains based on one or more chains of a multi-chain protein. During the ``chain-by-chain" generation, after completing a chain, we need to translate the generated parts. This is necessary because \method requires initialization at the origin when generating structures to meet the requirements of \texttt{SO(3)} invariance. Therefore, we need to move the generated parts away from the origin to make space for the next chain to be generated (we also tried not translating the parts, but found that the subsequently generated chains tended to wrap uniformly around the preceding chains).

The location where the generated parts are moved to determines where the binding interfaces of the complex appear, and this process can be manually specified. To avoid bias introduced by the manual specification of binding interfaces, we randomly select an amino acid on the generated part to serve as the binding site (we calculated the distance of each amino acid to the protein's center, then sorted them based on these distances, then we randomly selected from amino acids whose distances are ranked between the 33rd and 66th percentile). Subsequently, we translate the generated parts until the coordinate of the chosen binding site amino acid is at the origin, and continue to translate a small distance with the same direction (default 1\AA). \method then begins generating the next chain.

\begin{table}[h]
    \vspace{0mm}
    \centering
    \fontsize{9}{9}\selectfont
    \tabcolsep 10pt
    \renewcommand{\arraystretch}{1.0}
    \caption{$\Delta\text{G}$ and the \metric{RMSD} before and after all-atom relaxation for generated multi-chain proteins in ``chain-by-chain" manner. Both of the two metrics show significant differences.}
    \label{tab:multi_chain_uncond_cbc}
    \begin{tabular}{llll}
\toprule
\gthl Length  & $\Delta\text{G}_\text{RSC}$    & $\Delta\text{G}_\text{RAA}$   & RMSD      \\
\midrule
\gthl 50-100  & 312.66/-14.05  & -54.46/-54.45 & 1.64/1.44 \\
\gthl 100-100 & 235.52/-13.77  & -31.79/-26.71 & 2.05/1.70 \\
\gthl 100-200 & 1073.51/-11.01 & -40.29/-46.05 & 2.47/1.87 \\
\gthl 50-200  & 1012.53/-6.84  & -10.96/-50.67 & 2.79/2.12 \\
\gthl 100-300 & 323.44/-10.94  & -58.24/-53.81 & 2.42/2.00 \\
\bottomrule
\end{tabular}
    \vspace{2mm}
\end{table}

\method shows significant differences in generating multi-chain proteins in the chain-by-chain manner compared to generating all chains simultaneously.
As shown in \cref{tab:multi_chain_uncond_cbc}, there are three main differences observed in multi-chain proteins generated in the chain-by-chain manner:

\begin{compactitem}
\item The binding strength between chains shows a considerable decrease;
\item The structural differences before and after relaxation become more pronounced;
\item Without performing relaxation, structural clashes between chains are observed.
\end{compactitem}

These variations may arise due to several reasons: 

\begin{compactitem}
\item In the chain-by-chain manner, the chains are more independent from another chain, which leads to reduced binding strength;
\item The binding interface also impacts binding strength, and a randomly chosen binding interface may not be optimal;
\item Once a chain is generated, its structure remains unchanged, which might result in some local structural clashes that the model cannot entirely resolve. 
\end{compactitem}

\begin{figure*}
\begin{center}
\centerline{\includegraphics[width=1.0\textwidth]{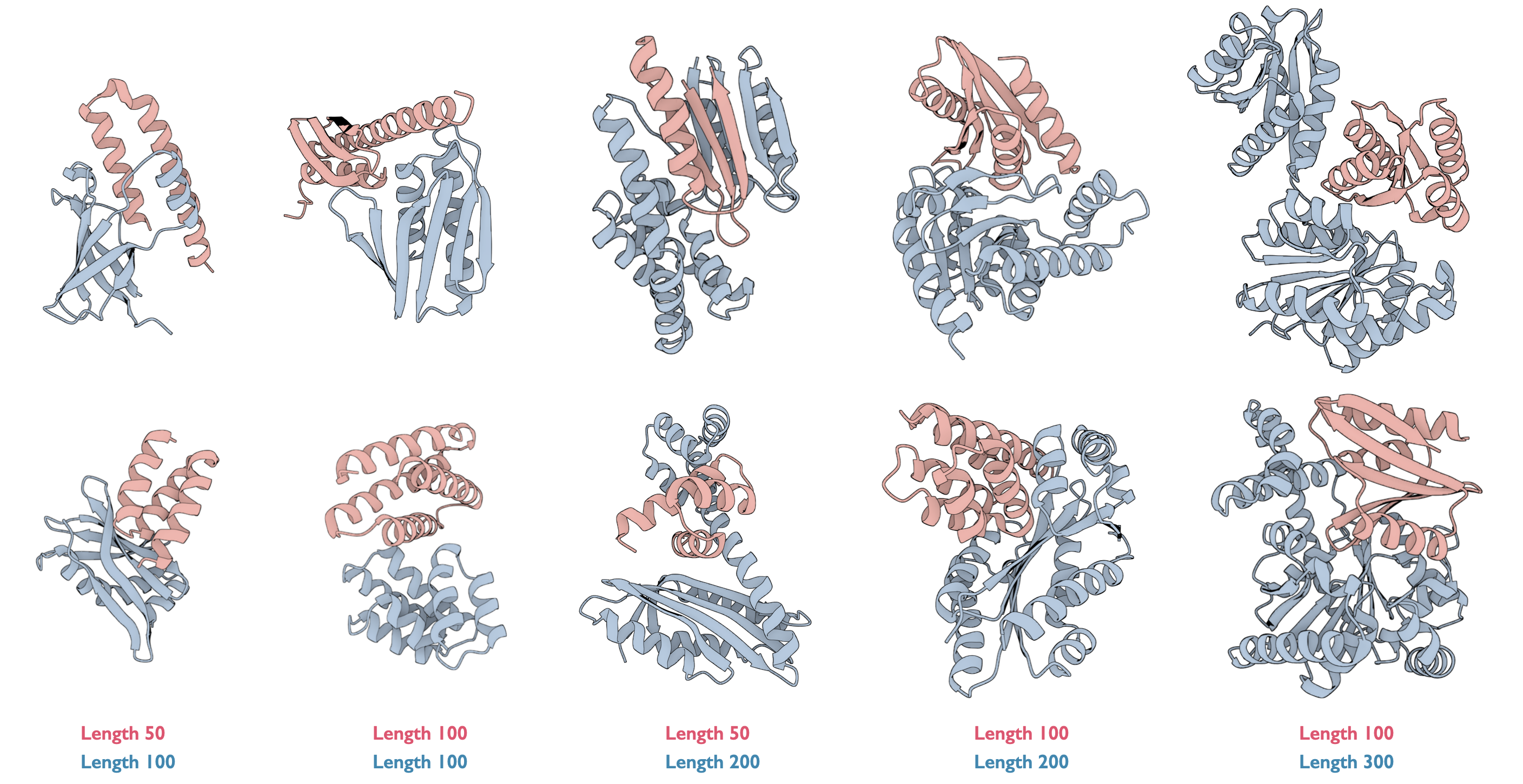}}
\caption{Samples generated by \method in the ``chain-by-chain'' manner. The generation order is from short chain to long chain, the length of each chain is highlighted with a unique color.}
\label{fig:multi_chain_cbc}
\end{center}
\end{figure*}

The samples generated in in chain-by-chain manner are shown in \cref{fig:multi_chain_cbc}.

\subsection{Antibody Design}
\label{appendix:experimental_details:antibody}

\paragraph{Data.}
Following~\citep{ProteinBench,AbDPO}, we use the Structural Antibody Database~\citep{SAbDab} under IMGT~\citep{IMGT} scheme as the dataset. 
We collected antigen-antibody complexes with both heavy and light chains and protein antigens and discarded the duplicate data with the same CDR-L3 and CDR-H3 sequence. 
The remaining complexes are used to cluster via MMseqs2~\citep{MMseqs2} with 40\% sequence similarity as the threshold based on the CDR-H3 sequence of each complex.
We then select the clusters that do not contain complexes in RAbD benchmark~\citep{RAbD} and split the complexes into training and validation sets with a ratio of 9:1 (1786 and 193 complexes respectively). 
The test set consists of 55 eligible complexes from the RAbD benchmark.

\paragraph{Methods.}
We follow the evaluation pipeline and results in ~\citet{ProteinBench}. 
We generate 64 candidates for each antigen. 
As our method directly generates all-atom structures, no additional sidechain packing tools are required.

\paragraph{Evaluation.}
Besides the traditional metrics like AAR and RMSD, we mainly focus on the quality in terms of generated CDR-H3's rationality and functionality towards the specific antigens. 
We utilize pyRosetta to perform the sidechain-only relaxation and calculate energy terms. 
Baseline methods only design the backbone structure, while energy calculations require the all-atom structure. Therefore, sidechain packing is indispensable. To avoid performance bias introduced by packing methods and conduct a fair comparison, we keep backbone structures fixed while applying relaxation to sidechain only.



Additionally, we also followed the procedure in the actual experiments, where both sidechain and backbone undergo relaxation before energy calculation, even though the structure at this point differs from the one designed by the model. The results are shown in \cref{tab:antibody_relaxbb}, \method still achieves the best performance.

\renewcommand{\arraystretch}{1.0}
\begin{table*}[!h]
    \caption{Performance comparison of antibody design methods on RAbD benchmark. ($\uparrow$/$\downarrow$) indicate whether higher or lower values are better. The best results are shown in \textbf{bold}.
    }
    \label{tab:antibody_relaxbb}
    \centering
    \footnotesize
    \begin{tabular}{lrr}
\toprule
Method & $\text{E} \downarrow$ & $\Delta\text{G} \downarrow$ \\

\midrule
dyMEAN & 72.76 & 36.43 \\
DiffAb & 14.56 & 2.29 \\
\midrule
\chl $\method_\text{SFT}$ & \chl \textbf{-3.72} & \chl \textbf{-4.08} \\
\chl $\method_\text{zero-shot}$ & \chl 10.38 & \chl -1.77 \\
\bottomrule
\end{tabular}
\end{table*}

\begin{figure}[h]
\begin{center}
\centerline{\includegraphics[width=1.0\textwidth]{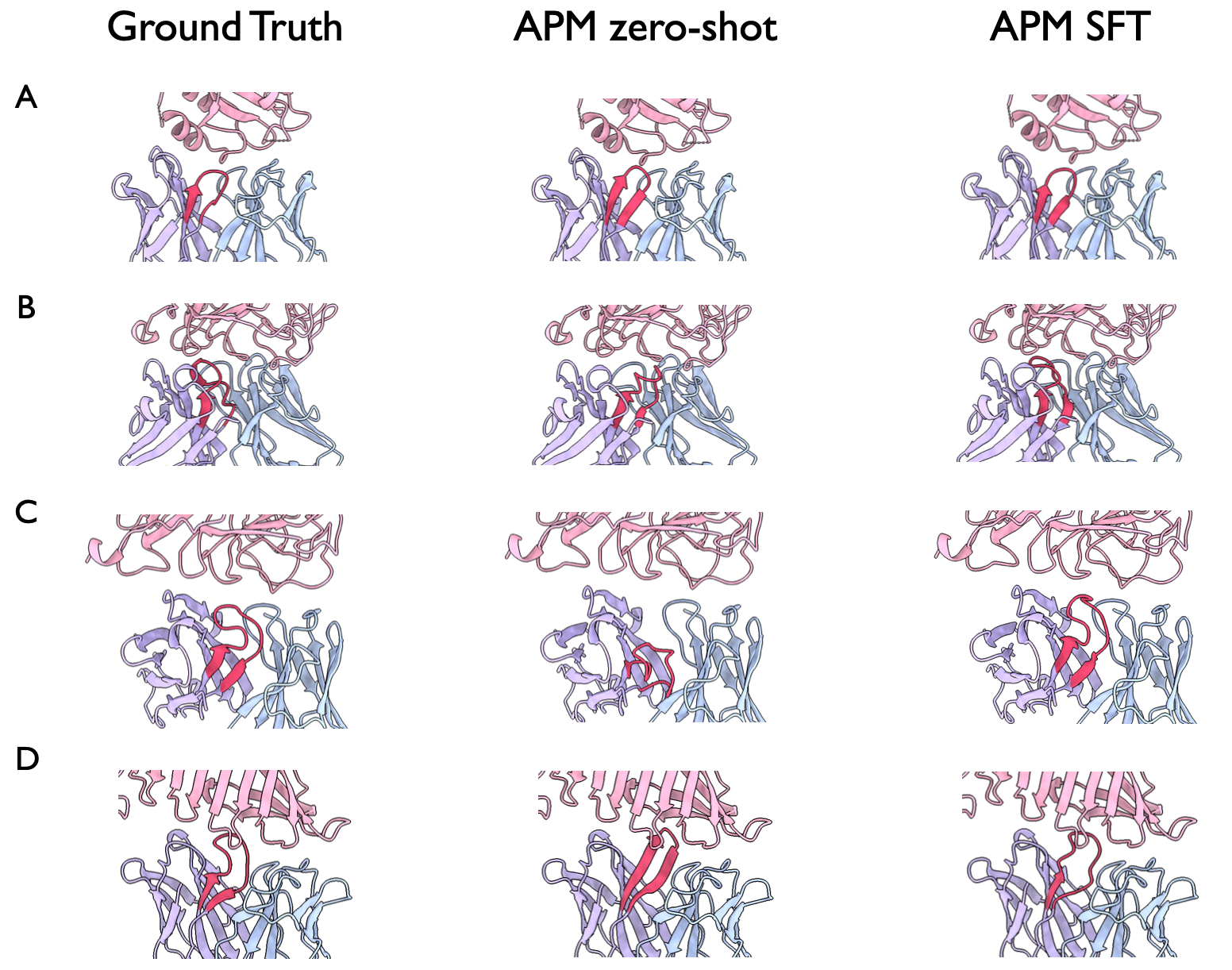}}
\vspace{2mm}
\caption{
The CDR-H3s generated in a zero-shot manner display distinct patterns compared to natural ones. The antigen is represented in {\color{pink}pink}, the antibody heavy chain in {\color{violet}purple}, and the antibody light chain in {\color{cyan}blue}, with the CDR-H3 highlighted in {\color{red}red}. The differences between CDR-H3 generated in a zero-shot manner and those from natural antibodies can be categorized as follows:
\textbf{A}. No obvious difference, the generated CDR-H3 closely resembles that of a natural antibody.
\textbf{B}. The generated CDR-H3 interacts with the antigen but binds at a different position compared to the reference antibody.
\textbf{C}. The generated CDR-H3 binds to the antibody light chain.
\textbf{D}. While the generated CDR-H3 binds correctly with the antigen, its structure predominantly consists of beta-sheets, unlike the 
common looped structures in natural antibody CDRs.
All the aforementioned differences disappear after undergoing SFT. The CDR-H3s generated by $\method_\text{SFT}$ closely resemble that of the natural ones.
}
\label{fig:antibody_zeroshot}
\vspace{0mm}
\end{center}
\end{figure}

\paragraph{Zero-shot.}
\label{appendix:experimental_details:antibody_zero-shot}
We observed that antibodies generated in a zero-shot manner exhibited stronger binding energy compared to those generated using the SFT model. 
However, there is a significant decrease in performance in terms of accuracy/similarity to natural antibodies. 
This phenomenon is expected since \method is trained to generate proteins capable of binding to other chains, allowing it to produce binding-capable antibodies without SFT.
However, because we excluded all antibody data from the training set, \method generates antibodies following the patterns of general proteins.
As a result, the generated CDR-H3, in both sequence and structure, differs from the natural ones, highlighting the differences between antibodies and general proteins.
Moreover, as we do not specify binding sites when designing proteins by using \method, the CDR-H3 designed without knowledge of antibody binding patterns might randomly bind antigens or antibody light chains.
To further illustrate this phenomenon, we selected some typical samples for visualization, as shown in the \cref{fig:antibody_zeroshot}.

\subsection{Peptide Design}
\label{appendix:experimental_details:peptide}

\paragraph{Data.}
The training and evaluation datasets are derived from PepBench~\citep{PepGLAD} and LNR~\citep{LNR}.
Following~\citet{PPFlow,PepGLAD}, we extract receptor pockets based on their spatial distances to the peptides. 
For data structure compatibility, we drop peptide-receptor samples that have non-standard amino acids during preprocessing.

\paragraph{Methods.}
For baselines, we follow the official open-source code for training and sampling.
We directly use the official checkpoint and scripts of PepGLAD\footnote{\url{https://github.com/THUNLP-MT/PepGLAD}} to sample candidates. 
For PPFlow and DiffPP, we carefully use their official code\footnote{\url{https://github.com/EDAPINENUT/ppflow}} for data preprocessing and follow their training instructions to obtain checkpoints at 200k steps. We also include RFDiffusion as a comparison method. Following the official guidelines\footnote{\url{https://github.com/RosettaCommons/RFdiffusion}}, we generate peptide structures using 50 diffusion timesteps and employ ProteinMPNN for sequence design. Additionally, we define the 6 amino acids on the receptor that are closest to the peptide as hot-spot residues, as we extract receptor pockets for other methods.

\paragraph{Evaluation.}
\label{appendix:experimental_details:peptide_evaluation}
Following ~\citet{PepGLAD}, we generate 40 peptide candidates for each sample using all methods.
Note that the length of each generated peptide is predefined to match its corresponding ground-truth sequence. 
We evaluate the binding capabilities of generated peptide candidates from multiple perspectives.

\begin{compactitem}
\item \textbf{Functionality.} We follow the approach used by ~\citet{PepGLAD}.
Both relaxation and energy calculations are performed using pyRosetta. 
The evaluation proceeds by identifying the best candidate for each receptor, and reporting the median performance values across all receptors.
Note that both the backbone and sidechain are applied relaxation. This procedure could achieve lower binding energies ($\Delta\text{G} < 0$) rather than the hundreds in antibody designing. 
It should be noted that this procedure may fail to achieve perfect relaxation in specific cases. For instance, 5 out of 93 ground truth samples still retained slight clashes, resulting in a $\Delta\text{G}$ slightly above 0(illustrated in \cref{fig:GT_dG_g0_samples}).

\begin{figure}[h]
\begin{center}
\centerline{\includegraphics[width=0.95\textwidth]{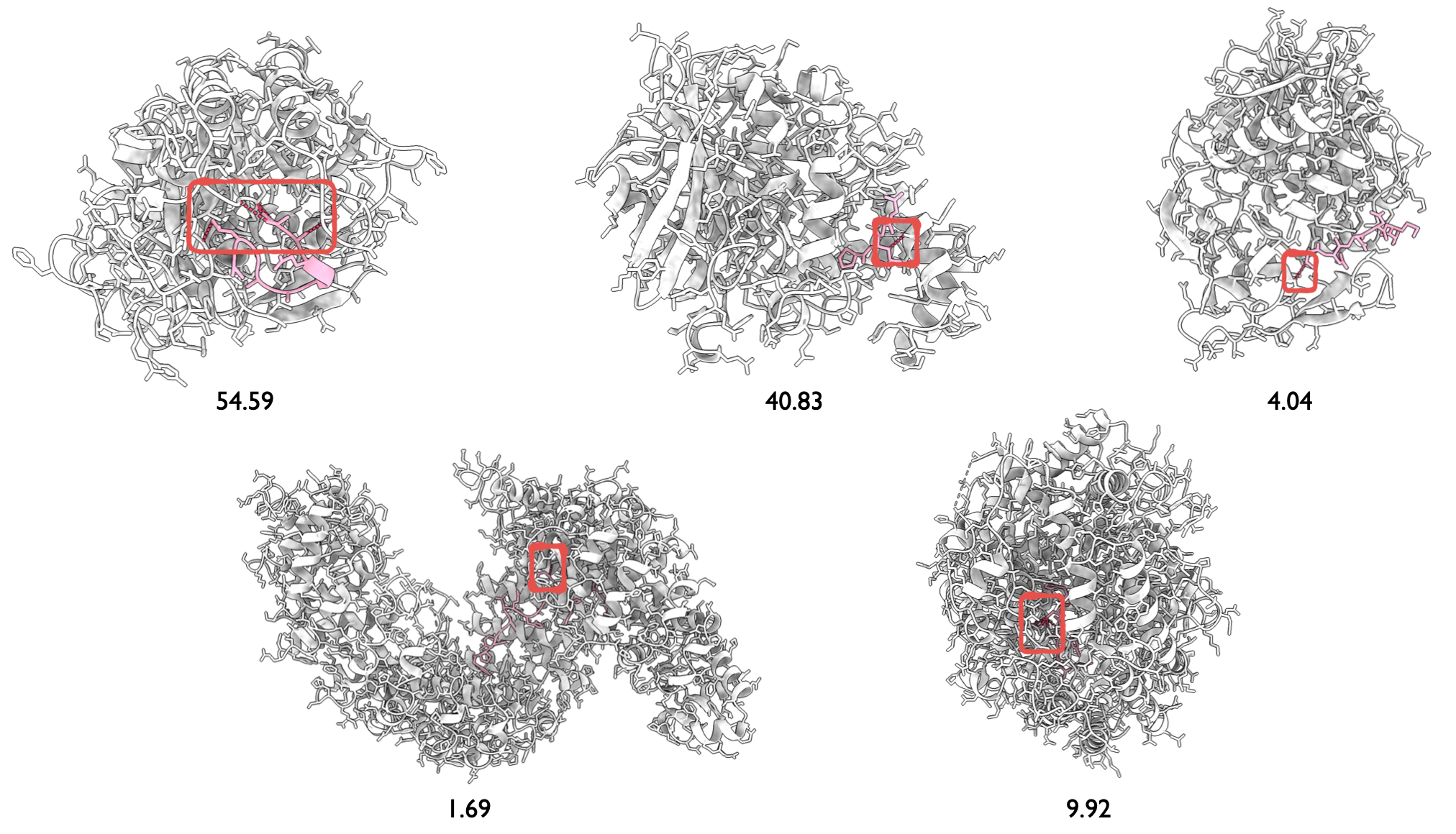}}
\vspace{-2mm}
\caption{5 ground truth samples with  $\Delta\text{G}$ exceeding 0. The receptor is shown in white, with the peptide highlighted in red. Slight clashes are marked by red boxes. Additionally, $\Delta\text{G}$ for each sample are listed below them.}
\label{fig:GT_dG_g0_samples}
\vspace{-2mm}
\end{center}
\end{figure}

\item \textbf{Foldability.} We only fold generated sequences with lengths greater than or equal to 10 residues (46 peptide-receptor pairs in total).
As indicated in the documentation of Boltz-1 and AlphaFold3, the \metric{ipTM} scores may not be reliable for sequences that are too short. 
To ensure reliability, we only consider peptide-receptor pairs where the ground-truth samples achieve successful confidence scores in Boltz-1 (\metric{pLDDT}$\ge 70$ and \metric{ipTM}$\ge 0.8$.), filtering out cases where even the ground truth structures fail to meet the confidence threshold.
The two threshold values are adopted from the official output documentation\footnote{\url{https://github.com/google-deepmind/alphafold3/blob/main/docs/output.md}} of AlphaFold3 and paper~\citep{AlphaFold3}.
Considering the computational cost and MSA retrieval time, we select the top 16 sequences ranked by $\Delta\text{G}$ and perform folding 8 times for each peptide-receptor pair. 
For each sequence, we select the best folding result, then average these results across all 16 sequences to obtain the final confidence score.
\item \textbf{Accuracy.} We measure the interface structural accuracy using the \metric{DockQ} score, which provides quality measures to quantify different aspects between generated and reference structures. 
Concretely, we utilize \metric{DockQv2}\footnote{\url{https://github.com/bjornwallner/DockQ}} from their official codebase.
Due to extreme structural conflicts observed in some samples generated by baseline methods, we only calculate \metric{DockQ} score for samples with $\Delta\text{G} \le 0$.
\end{compactitem}

\begin{figure*}[h]
\begin{center}
\centerline{\includegraphics[width=0.95\textwidth]{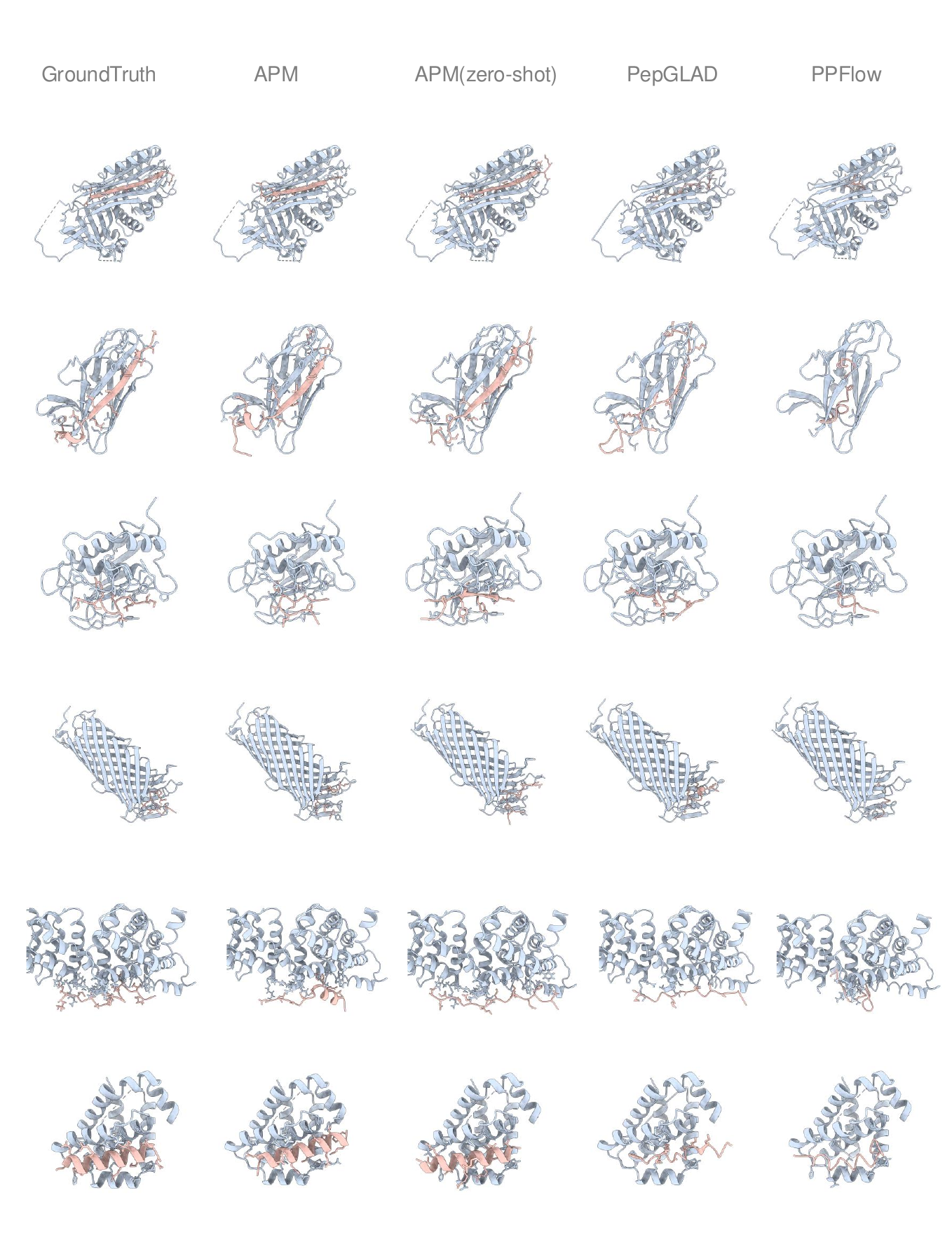}}
\vspace{-2mm}
\caption{Visualization of peptides generated by different methods. The {\color{cyan}blue} regions represent the given receptors, while the {\color{pink}pink} regions show the generated peptides, with all-atom structures displayed at the interface regions. From left to right: Ground truth structures, \method, \method zero-shot, PepGLAD, and PPFlow. The PDB IDs for the six cases from top to bottom are: \metric{1jrr}, \metric{2cnz}, \metric{3ayu}, \metric{4dcb}, \metric{5frs}, and \metric{6qg8}.}
\label{fig:peptide_raw}
\vspace{-2mm}
\end{center}
\end{figure*}

\begin{figure*}[h]
\begin{center}
\centerline{\includegraphics[width=0.95\textwidth]{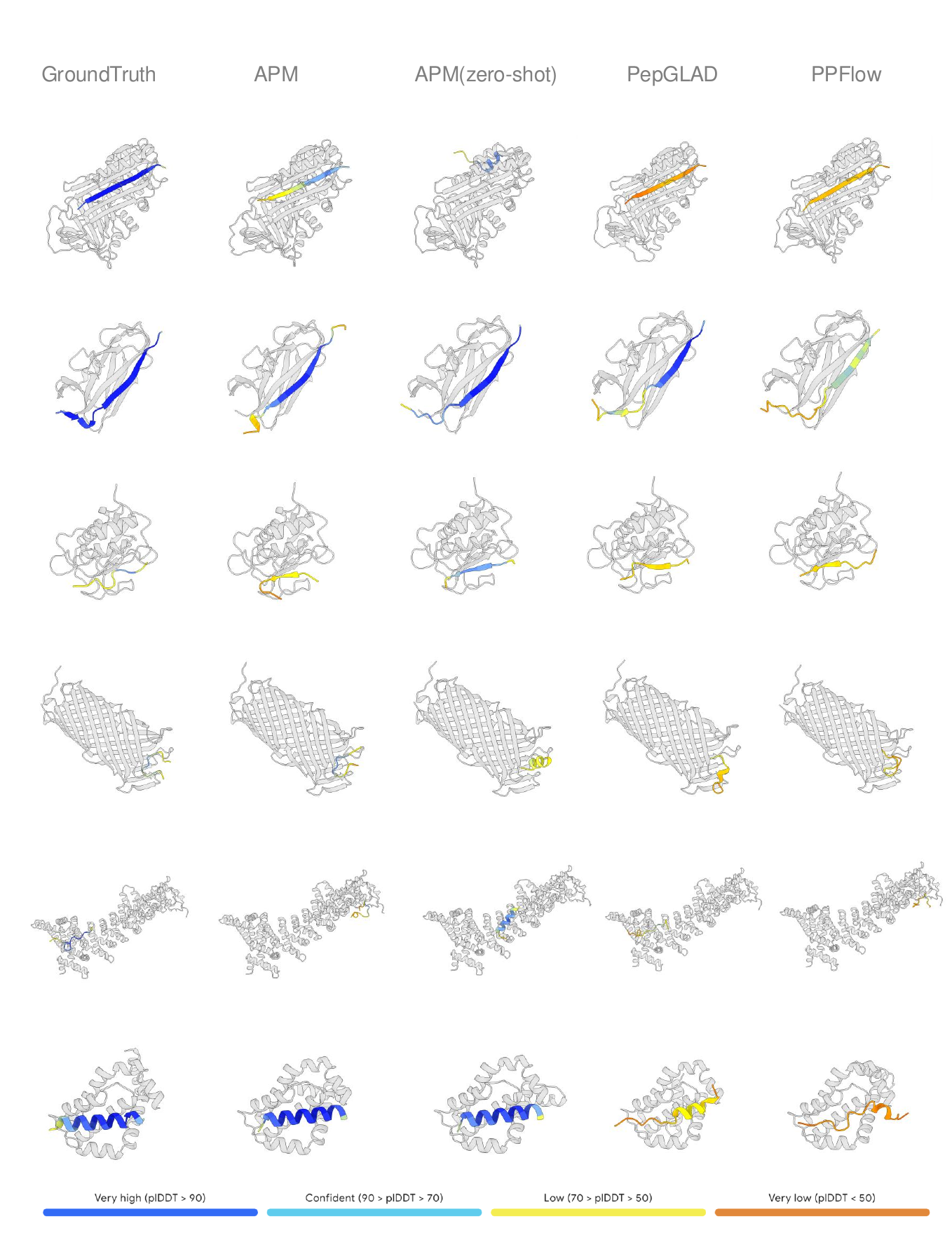}}
\vspace{-2mm}
\caption{Folded structures of sequences generated by different methods using Boltz-1. The structures are colored according to the AlphaFold-style pLDDT confidence scheme. From left to right: Ground truth structures, \method, \method zero-shot, PepGLAD, and PPFlow. The PDB IDs for the six cases from top to bottom are: \metric{1jrr}, \metric{2cnz}, \metric{3ayu}, \metric{4dcb}, \metric{5frs}, and \metric{6qg8}.}
\label{fig:peptide_plddt}
\vspace{-2mm}
\end{center}
\end{figure*}

\paragraph{Visualization} 
\begin{compactitem}
\item \textbf{Generated Structure.}
We present the peptides designed by different methods in \cref{fig:peptide_raw}.
The showcased examples include peptides with diverse secondary structures (loops, helices, and sheets). 
As observed in the visualization, our method demonstrates the ability to understand and generate appropriate secondary structures while maintaining reasonable interactions with the receptor.
\item \textbf{Folded Structure.}
As shown in \cref{fig:peptide_plddt}, we visualize the folded structures of sequences generated by different methods using Boltz-1. The structures are colored according to the pLDDT confidence, where blue regions indicate high confidence. 
We borrow the color bar from AlphaFold server website\footnote{\url{https://alphafoldserver.com/}}.
Receptors are shown in gray with 20\% transparency for better visualization. The visualization demonstrates that our method generates sequences capable of folding into stable structures with high confidence scores, indicate the quality of the generated sequences.
\end{compactitem}

\section{Extended Related Works}
\label{appendix:extended_related_works}

\paragraph{Sidechain Prediction.} 
Accurate prediction of sidechain conformation is crucial for protein design. 
Recent deep learning approaches have significantly advanced this field. 
DiffPack~\citep{DiffPack} employs a torsional diffusion model that learns the joint distribution of sidechain torsion angles by diffusing and denoising in torsional space. 
It autoregressively generates the four torsion angles. 
AttnPacker~\citep{AttnPacker} directly incorporates backbone 3D geometry to simultaneously compute all sidechain coordinates without relying on discrete rotamer libraries or conformation search.

\paragraph{Motif-Scaffolding.} 
Motif-scaffolding, the design of proteins that incorporate specific functional motifs emerged as a powerful approach in functional protein design.
Structure-based methods like RFDiffusion~\citep{RFdiffusion} and FrameFlow~\citep{FrameFlow} enable the generation of backbone scaffolds that can accommodate predefined motifs while maintaining overall structural stability. 
Sequence-based approaches including EvoDiff~\citep{EvoDiff}, DPLM~\citep{DPLM}, and ESM3~\citep{ESM3} present capabilities by designing sequences that fold into structures compatible with functional motifs. 
These methods collectively provide a comprehensive toolkit for designing proteins with specific functional properties.

\paragraph{Protein Structure Refinement.} 
Structure refinement is essential for optimizing protein designs to achieve native-like stability and function. 
Physics-based methods such as Rosetta relax~\citep{Rosetta} and OpenMM minimization~\citep{OpenMM} remain widely used for local refinement of protein structures. These refinement methods play a crucial role in the protein design pipeline, helping to bridge the gap between computational designs and experimentally viable proteins.

\section{Binder Design}
\label{appendix:binder_design}

\begin{figure}[tb]
\centering
\includegraphics[width=\textwidth]{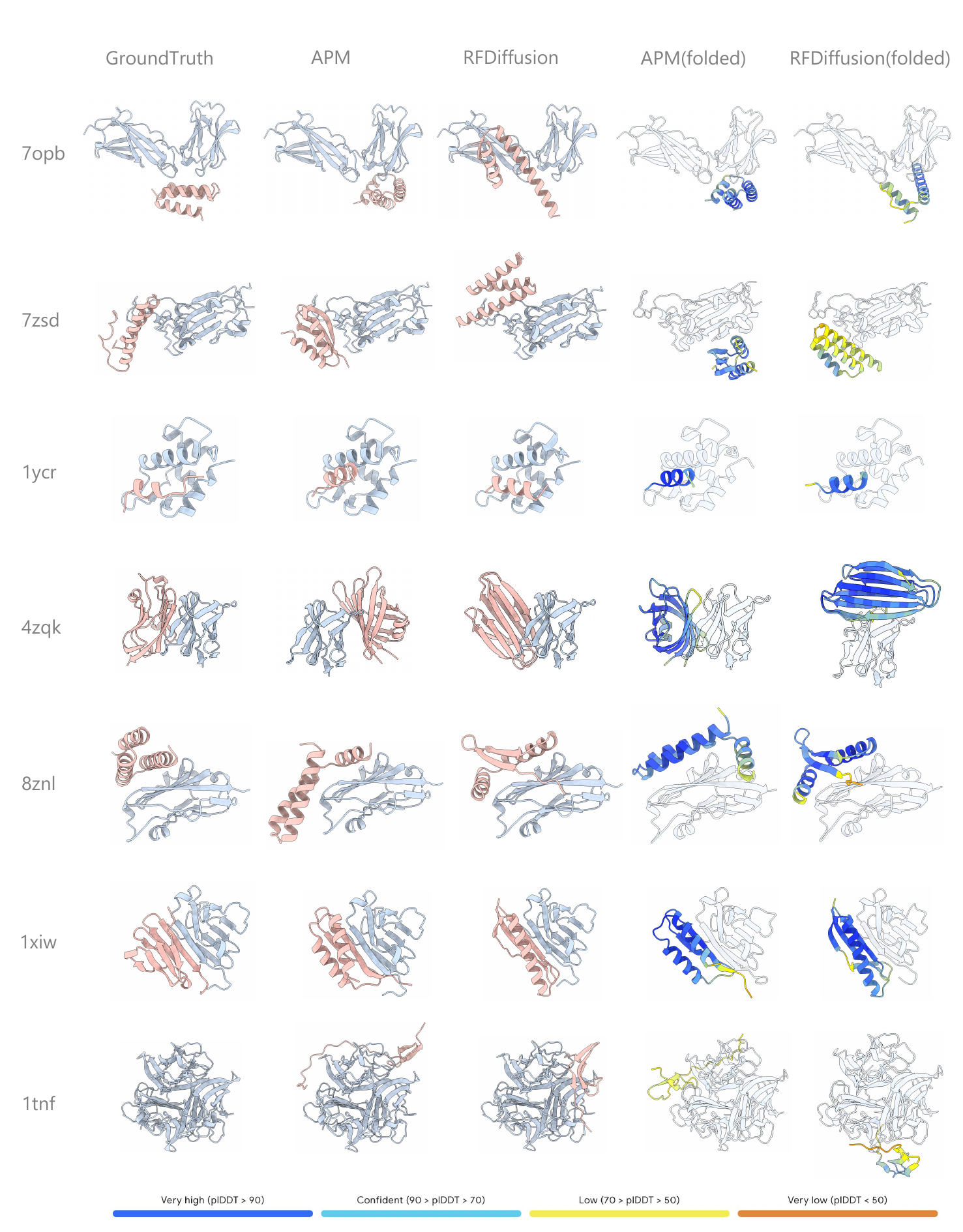}
\caption{Visualization of binder design results across six protein targets. From left to right: (1) GroundTruth: native complex structures, (2) APM: structures generated by APM, (3) RFDiffusion: structures generated by RFDiffusion, (4) APM(folded): highest pLDDT APM sequences folded with Boltz, (5) RFDiffusion(folded): highest pLDDT sequences folded with Boltz. The {\color{cyan}blue} regions represent the given targets, while the {\color{pink}pink} regions show the generated binders. For folded structures, targets are rendered transparent to highlight binders, with binders colored according to AlphaFold's pLDDT scheme. Each row represents a different PDB target: \metric{7opb}, \metric{7zsd}, \metric{1ycr}, \metric{4zqk}, \metric{8znl}, \metric{1xiw} and \metric{1tnf}.}
\label{fig:binder}
\end{figure}

\textbf{Settings. }
We further explore APM's zero-shot capabilities in binder design, focusing specifically on longer protein binders rather than short peptides. 
Following previous works~\cite{RFdiffusion,AlphaProteo}, we selected several important targets: Interleukin-7 Receptor-$\alpha$(IL-7RA), SARS-CoV-2 spike protein receptor-binding domain (SC2RBD), MDM2, Programmed Death-1 (PD1), Programmed Death-Ligand 1 (PD-L1), CD3-epsilon (CD3E), and TNF-$\alpha$. 
We extracted the corresponding target chains and reference binders from PDB: \metric{7opb}, \metric{7zsd}, \metric{1ycr}, \metric{4zqk}, \metric{8znl}, \metric{1xiw}, and \metric{1tnf} respectively.
The evaluation settings remain consistent with peptide design experiments. 
For the TNF-$\alpha$ target, the binder length was set to 50 residues, since no binding PDB structure was found.
However, due to uncertainty about which amino acids should be defined as hot-spot residues for these targets, we did not define hot-spot residues for RFDiffusion nor extract pockets for APM. 
Consequently, we do not report DockQ for this task. Additionally, due to computational resource constraints, we folded only the top 8 sequences rather than 16 as in previous experiments. 
The results of average metrics are presented in \cref{tab:binder}, with the 'Success' representing the proportion of samples (out of 8) that achieve both pLDDT $>$ 80 and ipTM $>$ 0.8.
We present the generated binders across different methods and their corresponding folded structures with the highest pLDDT sequences in \cref{fig:binder}.

\textbf{Discussion. }
Overall, APM is comparable to RFDiffusion, except in the case of the trimer target TNF-$\alpha$. Although these metrics have been proven by many studies to be predictive of wet lab experimental results~\cite{AlphaProteo,bennett2025atomically}, the actual effectiveness still requires validation through wet lab experiments.
We observed that both APM and RFdiffusion encounter cases where some samples exhibit lower pLDDT and ipTM scores. 
The pLDDT score can vary significantly along a protein chain. 
This means the folding model can be very confident in the structure of some regions of the protein, but less confident in other regions. 
We hypothesize that the low pLDDT scores stem from the complexity of long binders. 
Specifically, certain regions may be naturally highly flexible or intrinsically disordered, leading the folding model to assign low pLDDT scores to these residues (as indicated in ~\cite{guo2022alphafold2}).
Regarding ipTM, we speculate that the lower scores may result from the larger binding interfaces typical of long binders, which often involve multiple contact points or complex features such as convex or polar epitopes, or hydrophobic regions~\cite{AlphaProteo}. 
These structural complexities and biological properties can contribute to lower ipTM scores.

\textbf{Future Directions. }
As suggested in ~\cite{AlphaProteo,bennett2025atomically}, pLDDT and ipTM are predictive of binding success.
We would like to discuss potential approaches to improve long binder design.
APM was originally developed as a general-purpose model for complex modeling rather than a task-specific one, which presents challenges in the context of long binder design. 
This can be reframed as a question of how to adapt a general model into a domain-specialized one. 
Recent work ~\cite{bennett2025atomically}, provides valuable practical directions. 
The authors successfully transformed RFdiffusion into an antibody-specific model by fine-tuning it on antibody-antigen complex structures, demonstrating that domain-specific data can significantly enhance performance. 
Similarly, a feasible approach to enhance APM for long binder design would be to use a curated dataset of long binder-target complexes, potentially sourced from PDB or synthetic data.
Besides, post-training techniques offer another strategy to optimize the model for generating high-confidence designs. As demonstrated in ~\cite{AlphaProteo,bennett2025atomically}, pLDDT and ipTM correlate with binding success.
Building on this insight, we could implement preference optimization focused on these confidence metrics. 
Applying DPO-like~\cite{DPO,DiffusionDPO} algorithms, we can then train the model to favor high-confidence designs while avoiding low-confidence ones.

\renewcommand{\arraystretch}{1.0}
\begin{table*}[bt]
    \caption{Performance comparison of binder design. $\text{APM}_\text{MPNN}$  represents using ProteinMPNN to redesign sequences. }
    \label{tab:binder}
    \centering
    \footnotesize
    \begin{tabular}{lcccccc}
\toprule
Target & Method & $\Delta\text{G}$ $\downarrow$ & \%~$<$0 $\uparrow$ & pLDDT & ipTM & Success \\
\midrule
\multirow{4}{*}{7opb} & \gthl GroundTruth & \gthl -50.06 & \gthl - & \gthl 95.58 & \gthl 0.94 & \gthl - \\
 & RFDiffusion & -44.23 & 100\% & 71.14 & 0.55 & 25.00\% \\
 & $\text{APM}_\text{MPNN}$ & -50.25 & 100\% & 76.60 & 0.61 & 12.50\% \\
 & \chl APM & \chl -55.72 & \chl 100\% & \chl 73.73 & \chl 0.46 & \chl 12.50\% \\
\midrule
\multirow{4}{*}{7zsd} & \gthl GroundTruth & \gthl -36.65 & \gthl - & \gthl 81.26 & \gthl 0.25 & \gthl - \\
 & RFDiffusion & -43.99 & 100\% & 68.33 & 0.24 & 0\% \\
 & $\text{APM}_\text{MPNN}$ & -63.01 & 92.50\% & 62.34 & 0.38 & 0\% \\
 & \chl APM & \chl -48.10 & \chl 92.50\% & \chl 71.33 & \chl 0.39 & \chl 12.50\% \\
\midrule
\multirow{4}{*}{1ycr} & \gthl GroundTruth & \gthl -25.24 & \gthl - & \gthl 90.42 & \gthl 0.93 & \gthl - \\
 & RFDiffusion & -39.47 & 100\% & 78.49 & 0.81 & 25.0\% \\
 & $\text{APM}_\text{MPNN}$ & -33.27 & 90.00\% & 71.10 & 0.70 & 50\% \\
 & \chl APM & \chl -37.94 & \chl 90.00\% & \chl 66.28 & \chl 0.67 & \chl 25.0\% \\
\midrule
\multirow{4}{*}{4zqk} & \gthl GroundTruth & \gthl -39.36 & \gthl - & \gthl 94.03 & \gthl 0.87 & \gthl - \\
 & RFDiffusion & -29.35 & 87.50\% & 75.79 & 0.39 & 0\% \\
 & $\text{APM}_\text{MPNN}$ & -43.33 & 77.50\% & 79.10 & 0.36 & 0\% \\
 & \chl APM & \chl -45.27 & \chl 90.00\% & \chl 80.18 & \chl 0.39 & \chl 0\% \\
\midrule
\multirow{4}{*}{8znl} & \gthl GroundTruth & \gthl -44.24 & \gthl - & \gthl 95.54 & \gthl 0.95 & \gthl - \\
 & RFDiffusion & -30.20 & 97.50\% & 78.93 & 0.56 & 12.50\% \\
 & $\text{APM}_\text{MPNN}$ & -52.85 & 92.50\% & 78.00 & 0.59 & 12.50\% \\
 & \chl APM & \chl -44.14 & \chl 92.50\% & \chl 72.11 & \chl 0.47 & \chl 0\% \\
\midrule
\multirow{4}{*}{1xiw} & \gthl GroundTruth & \gthl -71.69 & \gthl - & \gthl 92.64 & \gthl 0.95 & \gthl - \\
 & RFDiffusion & -56.99 & 95.00\% & 77.22 & 0.76 & 62.5\% \\
 & $\text{APM}_\text{MPNN}$ & -46.96 & 82.50\% & 72.08 & 0.70 & 12.5\% \\
 & \chl APM & \chl -43.25 & \chl 85.00\% & \chl 73.27 & \chl 0.62 & \chl 12.5\% \\
\midrule
\multirow{4}{*}{1tnf} & \gthl GroundTruth & \gthl - & \gthl - & \gthl - & \gthl - & \gthl - \\
 & RFDiffusion & -74.35 & 97.50\% & 61.50 & 0.83 & 0\% \\
 & $\text{APM}_\text{MPNN}$ & -56.66 & 95.00\% & 62.52 & 0.83 & 0\% \\
 & \chl APM & \chl -48.99 & \chl 85.00\% & \chl 60.33 & \chl 0.84 & \chl 0\% \\
\bottomrule
\end{tabular}
\end{table*}

\section{Future Works}
\label{appendix:limitations}
\textbf{Model Scaling.} We chose not to incorporate triangular attention in \method, which is considered a key feature in the success of AlphaFold2/3, because we aim to scale the model in the future to observe whether scaling laws exist in our model. Triangular attention significantly restricts our ability to scale the model size.

\textbf{Pair Information from PLM.} In \method, PLM plays a crucial role by providing the model with a robust understanding of protein sequences. 
However, we only utilized the representations of individual amino acids from the PLM, and not the pair-level information (pair-level information refers to the attention matrix in PLMs). Pair-level information has been proven to provide significant benefits for protein structure learning. The reason we did not use pair-level information is to accelerate the encoding process of sequences by PLM (especially in representing multi-chain data). We used a PLM implemented with flash attention, which prevented us from obtaining complete pair-level information. In the future, we will attempt to resolve the encoding speed issue with PLM and use the original PLM implementation to gain the access to pair-level information.

\textbf{Refine Module.} The \rfm is designed to refine the structure and sequence generated by the \sbm, which can be seen as a form of relaxation that allows modifications to the types of amino acids. Currently, our \rfm primarily aims to make the generated proteins resemble real proteins more closely. In the future, we will attempt to incorporate more biological/physical constraints (e.g., force fields) into the \rfm to achieve better performance.

\textbf{Interchain Hotspot Residue Assignment.} Hotspot residues, the amino acids playing crucial roles in interactions, also be considered as regions where interactions occur, serving as important parameters in describing the binding proteins.
However, the current version of \method does not support specifying hotspot residues during the generation of complexes or functional proteins. 
Instead, \method autonomously determines the regions where interactions occur. 
This design was made to avoid the impact of assigning different hotspot residues on the model's performance during general multi-chain protein generation.
Consequently, this design leads to differences in the binding patterns of proteins generated directly using \method (in a zero-shot manner) compared to natural samples.
In the future, we will address this issue to allow \method to support the assignment of hotspot residues, thereby enhancing \method's performance in a zero-shot manner.

\textbf{Downstream Tasks.} We will validate \method’s capability in designing proteins with biological functions in more downstream tasks.

\section{Visualization}
All protein visualizations in this paper were completed using ChimeraX \citep{ChimeraX2023} (\cref{fig:interaction}, \cref{fig:illustration}, \cref{fig:training}, \cref{fig:antibody_zeroshot}, \cref{fig:GT_dG_g0_samples}, 
\cref{fig:peptide_raw}, \cref{fig:peptide_plddt},
\cref{fig:binder}) and Protein Viewer \citep{ProteinViewer} (the remaining visualizations).


\end{document}